\documentclass[lettersize,journal]{IEEEtran}
\usepackage{amsmath,amsfonts}
\usepackage{algorithmic}
\usepackage{algorithm}
\usepackage{array}
\usepackage[caption=false,font=normalsize,labelfont=sf,textfont=sf]{subfig}
\usepackage{textcomp}
\usepackage{stfloats}
\usepackage{url}
\usepackage{verbatim}
\usepackage{graphicx}
\usepackage{multirow}
\usepackage{booktabs}
\usepackage{caption}
\usepackage{tabu}
\usepackage{subcaption}
\usepackage{color}
\usepackage{xcolor}
\usepackage{cite}
\newcommand{\ie}{\textit{i}.\textit{e}.}
\newcommand{\eg}{\textit{e}.\textit{g}.}
\begin{document}

\title{Collaboratively Self-supervised Video Representation Learning for Action Recognition}

\author{
Jie Zhang,~\IEEEmembership{Member,~IEEE,}
Zhifan Wan,
Lanqing Hu,
Stephen Lin,~\IEEEmembership{Member,~IEEE,}
Shuzhe Wu,
Shiguang Shan,~\IEEEmembership{Fellow,~IEEE}
}

\markboth{Journal of \LaTeX\ Class Files,~Vol.~14, No.~1, July~2024 }%
{Shell \MakeLowercase{\textit{et al.}}: A Sample Article Using IEEEtran.cls for IEEE Journals}


\maketitle
\renewcommand{\thefootnote}{\fnsymbol{footnote}}
\footnotetext[0]{
Manuscript received 5 July 2024; 
revised 8 December 2024 and 1 January 2025; 
accepted 2 January 2025. 
Date of publication 16 January 2025; 
date of current version 15  January 2025.
This work is partially supported by Natural Science Foundation of China (No. U2336213, 62176251) , Strategic Priority Research Program of the Chinese Academy of Sciences (No. XDB0680202),
Beijing Nova Program (20230484368), Suzhou Frontier
Technology Research Project (No. SYG202325),
and Youth Innovation Promotion Association CAS.
The associate editor coordinating to the review of this manuscript amd approving it for publication was Prof. Zhen Lei. \textit{(Corresponding author: Shiguang Shan.)}

Jie Zhang, Zhifan Wan, Lanqing Hu and Shiguang Shan are with the Key Laboratory of AI Safety of CAS, Institute of Computing Technology (ICT), Chinese Academy of Sciences (CAS), Beijing 100190, China. 
They are also affiliated with the School of Computer Science and Technology, University of Chinese Academy of Sciences (UCAS), Beijing 100049, China (e-mail: zhangjie@ict.ac.cn; wanzhifan16@mails.ucas.ac.cn; lanqing.hu@vipl.ict.ac.cn; sgshan@ict.ac.cn).

Stephen Lin is with Microsoft Research Asia, Beijing 100080, China (e-mail: stevelin@microsoft.com).

Shuzhe Wu is with Beijing Huawei Digital Technologies Co., Ltd., No.3 Xinxi Road, Haidian District, Beijing 100095, China (e-mail: wushuzhe2@huawei.com).
}
\renewcommand{\thefootnote}{\arabic{footnote}} 

\begin{abstract}
Considering the close connection between action recognition and human pose estimation, we design a Collaboratively Self-supervised Video Representation (CSVR) learning framework specific to action recognition by jointly factoring in generative pose prediction and discriminative context matching as pretext tasks. Specifically, our CSVR consists of three branches: a generative pose prediction branch, a discriminative context matching branch, and a video generating branch. Among them, the first one encodes dynamic motion feature by utilizing Conditional-GAN to predict the human poses of future frames, and the second branch extracts static context features by contrasting positive and negative video feature and I-frame feature pairs. The third branch is designed to generate both current and future video frames, for the purpose of collaboratively improving dynamic motion features and static context features. Extensive experiments demonstrate that our method achieves state-of-the-art performance on multiple popular video datasets.
\end{abstract}


\section{Introduction}

Human action recognition is a fundamental task in computer vision and has gained increasing attention in recent years. With a large number of labeled videos, the convolutional neural network (CNN) has achieved great success in human action recognition by learning powerful spatio-temporal representation features \cite{feichtenhofer2020x3d,feichtenhofer2019slowfast,tran2019video,tran2018closer}. However, the cost of collecting and tagging large amounts of videos is enormous. Fortunately, there are a large amount of unlabeled videos on the Internet. A common practice is to use a large amount of unsupervised video for robust feature representation learning and then transfer to downstream tasks such as human action recognition, which deserves further exploration.

Self-supervised learning has emerged as a promising approach for unsupervised learning. Its goal is to learn powerful feature representation by designing appropriate pretext tasks which automatically generate free annotations for unlabeled data and transfer the representations to various downstream tasks. This approach has achieved remarkable success in areas such as image recognition, object detection, and image segmentation.
In addition to self-supervised learning on images \cite{caron2018deep,chen2020simple}, there has been a growing interest in exploring self-supervised learning for video representation with the increasing availability of large-scale unlabeled video data.

Early self-supervised video learning approaches focus mainly on effectively exploiting contextual relationships within videos \cite{huo2020self,wang2020self,benaim2020speednet}. 
Over the years, self-supervised methods have become more powerful and can generally be categorized into two groups: discriminative classification / regression methods \cite{qian2021spatiotemporal,guo2022cross} and generative prediction methods \cite{lin2024tsgan,dsm2020, feichtenhofer2022masked}. Discriminative methods focus on learning to distinguish between different samples, while generative methods aim to predict missing information. 
Recent efforts have been made to learn general video representations that can be applied to a range of downstream tasks. This includes video contrastive learning \cite{qian2021spatiotemporal, jenni2021time, dave2022tclr,ni2022motion,feichtenhofer2021large}, as well as cross-modal learning \cite{videobert2019, gdt2020}, which learns from different modalities such as video, audio, and text.
Although these methods achieve promising results for various video based downstream tasks, they do not explicitly consider designing pretext tasks for action understanding, which may lead to sub-optimal solutions for action recognition. Considering the close connection between action recognition and human pose estimation, we design a self-supervised video representation learning framework specific to action recognition by explicitly inducing generative pose prediction as the pretext task. One interesting work, DCM \cite{huang2021self} designs a two-stream network to separately learn action-related features, i.e., context and motion features, which achieves impressive results for action recognition. However, DCM fails to jointly consider the learning of context and motion features, lacking comprehensive view of these two features. Differently, we conduct collaboratively self-supervised video feature learning by jointly considering generative pose prediction and discriminative context matching as pretext tasks.

\begin{figure*}[!tb]
  \centering
  \includegraphics[width=1\textwidth]{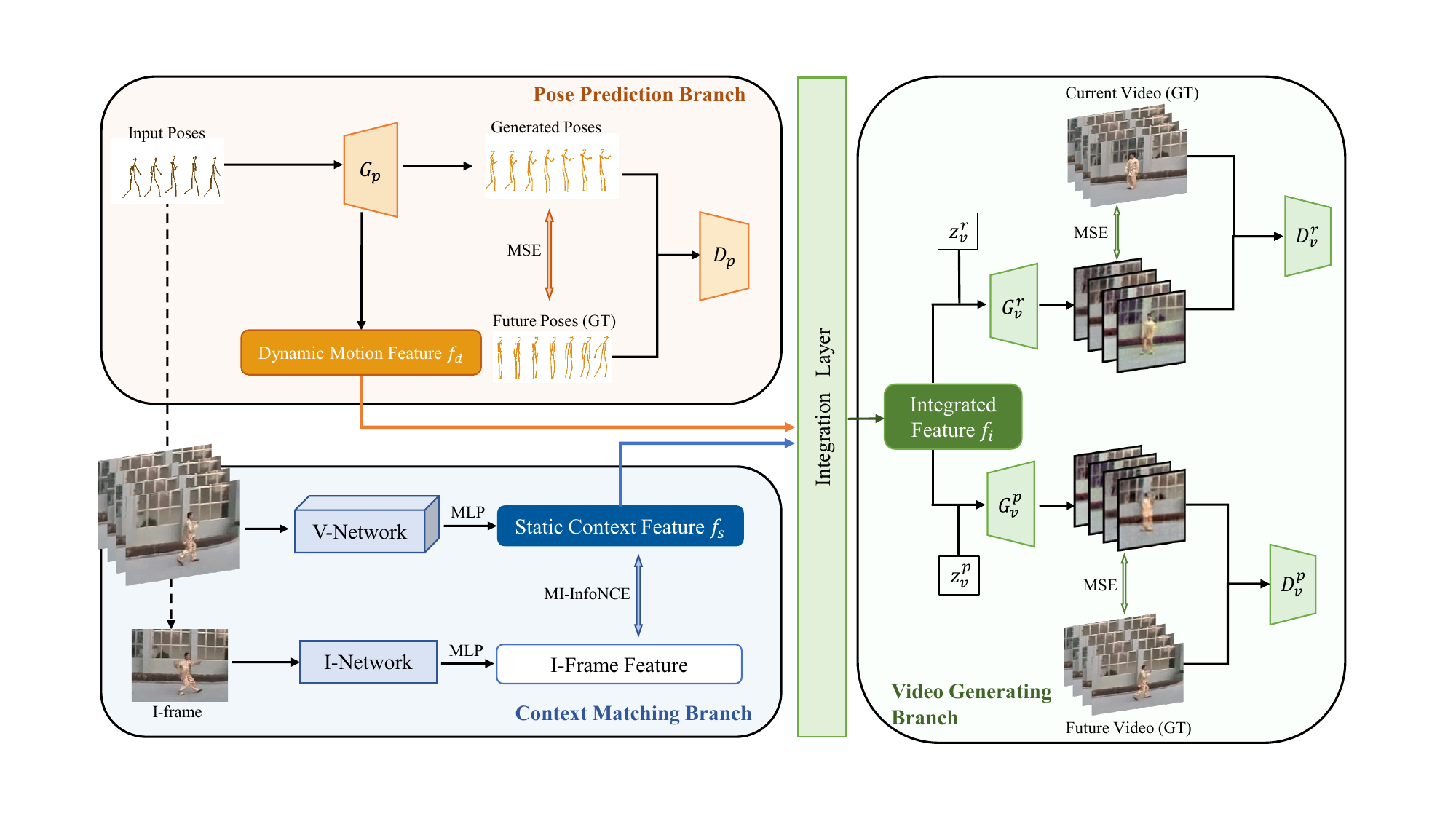}
  \vspace{-1cm}
  \caption{An overview of our self-supervised learning progress. 
  Our CSVR contains three branches: pose prediction branch, context matching branch, and video generating branch. 
  The pose prediction branch uses poses from input video clips to predict future poses. Dynamic motion feature $f_d$ is originating from the encoder of pose generator.
  The context matching branch captures video clip features and intra frame (I-frame) features by CNNs, and then casts a contrastive loss on learned features to pull the feature pairs from the same video clips together. Through the contrastive learning method, video clip feature $f_s$ can contain rich static context information.
  After fusing previous features by the integration layer, integrated feature $f_i$ is fed into the video generating networks, try to reconstruct current 
 video clips and predict future video clips, respectively. The video generating branch jointly optimizes all the branches, leading to learn more comprehensive video representation. 
  }
  \label{fig:pipline}
\end{figure*}

In this work, we propose the \textbf{C}ollaboratively \textbf{S}elf-supervised \textbf{V}ideo \textbf{R}epresentation (CSVR) framework, which jointly learns dynamic motion and static context features for the downstream action recognition task.
As illustrated in Figure \ref{fig:pipline}, CSVR consists of three main representation learning branches, \ie pose prediction branch, context matching branch, and a collaborative training branch named video generating branch. 
To enhance the performance of action recognition, we resort to the strong relationship between human pose and actions. Specifically, we employ a generative pose prediction branch based on a conditional generative adversarial network (CGAN) \cite{mirza2014conditional}, which predicts future poses by taking current pose sequences as input. By excluding the potential motion noise from backgrounds, our method can effectively learn dynamic motion features highly related to action.
Intra frame (I-frame) refers to a type of frame in video compression that contains complete image data and can be used as a reference point for decoding other frames.
It is typically considered to to contain valuable context information.
To leverage the abundant pattern in I-frame for action recognition, we incorporate a discriminative context matching branch, which works by pulling the features of clips and I-frames from the same videos closer together, while simultaneously pushing apart pairs from different videos.
Finally, we design the collaborative video generating branch to jointly optimize the dynamic and static features. The branch contains two video generating networks, one aims to reconstruct current video clips, another is trained for predicting future video clips. 
The video generation process requires both static context information and dynamic motion information, which provides a powerful collaborative learning objective, assisting the previous two branches in learning excellent static and dynamic features.

Extensive experiments on four popular datasets with various backbones show that our CSVR achieves state-of-the-art performance for two downstream action understanding tasks, \ie action recognition and video retrieval.

We summarize the contributions in the following:

1) Considering the strong connections between human pose and actions, we introduce a novel framework focusing on action recognition named Collaboratively Self-supervised Video Representation learning (CSVR), which simultaneously learns dynamic motion and static context features using three branches with discriminative and generative pretext tasks.
    
2) Our carefully designed collaborative video generating branch enables joint optimization of dynamic motion and static context features, resulting in more comprehensive representations that benefit downstream action understanding tasks.

\section{Related Work}\label{sec2}

As our method is primarily focused on self-supervised video representation learning for action recognition, in this section, we provide a brief overview of some typical works in the fields of video action recognition and self-supervised video representation learning. 
Additionally, we review some generation methods as GAN-based \cite{goodfellow2020generative} generative branches play a crucial role in our method.

\subsection{Video-Based Action Recognition}

Videos contain motion information with a complex temporal structure, which requires spatio-temporal feature learning. To address this, numerous advanced 3D-CNN architectures have been proposed \cite{feichtenhofer2020x3d, feichtenhofer2019slowfast, resnet2p1d_2018, s3d2018}. 
Recently, with the inspiration of Transformer\cite{transformer2017} architectures in natural language processing, researchers have attempted to extend Vision Transformer (ViT) \cite{dosovitskiy2020image} for video modeling. Many works \cite{arnab2021vivit,bertasius2021space,bulat2021space} propose different variants for spatiotemporal representation learning, and adapt to video action recognition task.
However, learning favorable motion representations from original video clips remains a challenging task. To this end, various movement information have been previously examined to help extracting human action information from raw video, such as optical flow \cite{coskun2022goca}, motion vectors \cite{coviar2018,dmc2019}, and pose data \cite{schneider2022pose,rai2021cocon}. With the advancement of body pose estimation methods \cite{cao2017realtime}, several approaches have utilized pose sequence information for assisting video action recognition. 
Previous works directly encode temporal correlation within pose sequences for action understanding \cite{liu2016spatio,shi2019two,caetano2019skeleton}.  
The high efficiency of video networks also leads to breakthroughs in self-supervised video representation learning for action recognition \cite{huang2021self,coskun2022goca,das2021vpn++, das2020vpn}. 
Some of them \cite{das2021vpn++, das2020vpn} use pose information as an attention or supervision signal to mine video representations more suitable for action understanding tasks.

\subsection{Self-Supervised Video Representation Learning}

Self-supervised video representation learning methods aims to leverage unlabeled video data to learn representations that can be easily transferred to downstream tasks. 
Existing methods for self-supervised video representation learning can be generally categorized into three types: context-based methods, contrastive learning and generative algorithms\cite{gui2023survey}. 
Early context-based methods focus on designing pretext tasks that exploit contextual relationships within videos, such as speed prediction \cite{huang2021self}, pace prediction \cite{wang2020self}, and jigsaw solving \cite{huo2020self}.
Contrastive learning methods focus on how to construct positive and negative examples without annotations and pull positive examples closer to each other than negative ones in training process \cite{behrmann2021long, pacepred2020, qian2021enhancing, caron2021emerging, chenempirical}. Interestingly, in contrast to the CNN-based approaches, there are a few very recent studies \cite{caron2021emerging, chenempirical} leveraging contrastive learning for transformers.
Recently, there has been a surge of research into Transformer-based generative mask video modeling methods \cite{feichtenhofer2022masked, tong2022videomae, wei2022masked}. These works learn representations by exploiting co-occurrence relationships between video patches in reconstructing process. 
Moreover, \cite{wei2022masked} tries to reconstruct features from multiple sources to perform self-supervised pretraining in videos.

Although self-supervised video representation learning methods achieve impressive performance on various downstream tasks including action recognition, most of them are not specially designed for action-oriented representation learning.
Differently, DCM \cite{huang2021self} and P-HLVC \cite{schneider2022pose} aims to learn generalizable features specific to action understanding. They simply extract motion features by contrastive learning and leverage diverse motion information, \ie pose and motion vectors as self-supervised signals. 
Inspired by these approaches, we propose a Collaboratively Self-supervised Video Representation (CSVR) learning framework that simultaneously considers pose prediction and context matching as pretext tasks. By jointly optimizing the whole framework by video generating branch, CSVR provides a more comprehensive feature representation for downstream action recognition task.

\subsection{Video Generation}
Over the past years, many efforts are devoted to video generation, which can be generally categorized into three groups: Generative Adversarial Networks (GAN) \cite{goodfellow2020generative}, Variational Autoencoders (VAE) \cite{kingma2013vae} and Diffusion Models \cite{ho2020diffusion}.
Generative adversarial network (GAN) \cite{goodfellow2020generative} is initially used to generate high-fidelity samples that matched the data distribution. Some work\cite{pan2021videomoco,lin2024tsgan} also employed it as a pretext task for video representation learning.
GANs usually consist of two encoder-decoder architecture networks: a generator producing fake data and a discriminator distinguishing between fake and real data.  
However, the inherent randomness of GAN output makes it difficult to control the generated data. To address this issue, Conditional GANs (CGAN) \cite{mirza2014conditional} are proposed for incorporating input conditions such as class labels to regulate the synthetic outputs. 
Recently, more efforts are devoted to video generation tasks. TGAN v2\cite{saito2020tganv2} is a video generation network that can efficiently generate realistic videos. 
Some works \cite{pan2021videomoco,liang2017dual,ghadekar2023semi,lin2024tsgan} prove the feasibility of learning video representation from generative learning methods. 
VideoMoCo\cite{pan2021videomoco} is the first work to use GANs for video representation learning. It reconstructs the foreground and background of videos separately using two GANs and applies the encoder of generator to action recognition downstream tasks. 
TS-GAN\cite{lin2024tsgan} employs a spatio-temporal two-stream GAN network to reconstruct keyframes and optical flow information, leveraging the learned features in downstream tasks. 
Unfortunately, both approaches underperformed across multiple datasets.
Although these methods using GANs can effectively reconstruct video dynamics, their features do not perform well in downstream classification tasks. 

Variational Autoencoders (VAEs) \cite{kingma2013vae} have become widely adopted in two-stage generative models, where the first stage compresses images into a lower-dimensional latent representation, and the second stage generates images from this latent space. In the context of video generation, 2D VAEs are often extended to 3D variants by incorporating 3D convolutional layers or temporal attention mechanisms \cite{ge2022long,villegas2022phenaki,yan2021videogpt,zhao2024cv}. These approaches have demonstrated promising results in video generation. More recently, diffusion models have achieved remarkable success in both image \cite{nichol2021improved,ramesh2022hierarchical,rombach2022high,saharia2022photorealistic} and video generation \cite{ho2022video,voleti2022mcvd,blattmann2023align,ge2023preserve,singer2022make,luo2023videofusion,gupta2025photorealistic}. Video diffusion models can be broadly categorized into pixel-space methods \cite{ho2022video,singer2022make} and latent-space methods \cite{blattmann2023align,ge2023preserve}. Compared to the pixel-space methods, latent-space methods offer significant efficiency advantages for video generation. However, they still incur high computational costs, which makes them challenging to apply in self-supervised video representation learning.
Therefore, in our CSVR, we employ 1D-CNN-based CGAN in pose prediction branch and R3D-based TGAN v2 in video generating branch. To enhance the stability and efficiency of our training process, we also adopt the gradient penalty from WGAN-GP \cite{gulrajani2017improved} and the parameter update strategy from OSGANs \cite{shen2021osgan}. These carefully-designed objectives help our method efficiently generate more accurate and diverse samples, thus improve the quality of learned dynamic motion representations.

\section{Methodology}
\label{sec:methodology}
\subsection{Overall Framework}
In this paper, we propose a novel framework for action recognition called Collaboratively Self-Supervised Video Representation learning (CSVR). The framework aims to learn robust and comprehensive feature representation for action recognition, which consists of both dynamic motion and static context features. To achieve this, we design three branches: a generative pose prediction branch for learning dynamic motion information, a discriminative context matching branch for learning static context representation, and a collaborative video generating branch to jointly optimize all branches. Figure \ref{fig:pipline} provides an overview of our method.

Human pose modeling is an intuitive and effective way to represent human motion. Therefore, how to effectively learn the dynamic information contained in poses has become the core issue for human action analysis. 
Previous work has made great progress in action recognition tasks using optical flow \cite{han2020self} and motion vectors \cite{huang2021self}. 
However, these approaches may perform poorly in practical scenarios due to noise information, such as irrelevant background movement and low-quality video noise, which can be naturally relieved by extracting poses from raw videos. 
Although some works \cite{schneider2022pose} try to learn dynamic information from poses by contrastive learning approaches, which may fail to fully leverage temporal information in the pose sequences, leading to unsatisfactory results in action recognition task.
In CSVR, we resort to designing a generative pose prediction branch to capture dynamic information. 
Specifically, we use OpenPose \cite{cao2017realtime} to estimate poses from raw video clips\footnote{Although we use the off-the-shelf pose estimator OpenPose, which is trained with pose supervision, we provide no additional supervision for the action recognition tasks. It does not violate the motivation of the self-supervised learning.}, and then deploy CGAN to predict future poses from the current ones. To make the training process even more concise, we utilize time-efficient unidimensional convolutional neural networks (1D-CNNs), which output pose sequences all at once. 

Besides the dynamic information from human poses, the context information is also important for action recognition. For the video compression, context information is roughly separated into intra frames (I-frame), which contain the main content of the video clip.  Thus, we conduct a context matching branch that compares global features from video clips with static image features from I-frames to learn static context information. Specifically, we randomly select video clips and extract I-frames following the practice of \cite{wu2018compressed}. We then utilize the MultiInstance-InfoNCE (MI-InfoNCE) loss \cite{diba2021vi2clr} to learn static context features by matching the video and I-frames. This branch pulls together features of videos and I-frames from the same video clips while pushing apart pairs from different clips. 
By matching still images to video clips, the learned representation is supposed to capture the contextual information beneficial for downstream action recognition task.

Since the pose prediction and context matching branches just learn dynamic and static features separately, it is significant to effectively integrate them to obtain comprehensive representations for downstream tasks. The video generating branch includes an integration layer and two video generation networks that provide powerful target for collaborative learning. The integration layer combines the dynamic and static features by Adaptive Instance Normalization (AdaIN) \cite{huang2017arbitrary}. We then feed the integrated features into TGAN-based \cite{saito2020tganv2}  video generation networks for two parallel tasks, video reconstruction and video prediction. 

Our collaborative training framework leverages the strong correlation between human action and poses, as well as between context and I-frames. By jointly learning dynamic motion and static context features with three branches, rich representations can be achieved that contributes to action-related downstream tasks, such as action recognition. We will introduce the detail of each branch in the following sections.

\subsection{Generative Pose Prediction Branch}
The generative pose prediction branch is conducted to learn dynamic motion features, which consists of two modules, \ie a pose generator ($G_p$) and a pose discriminator ($D_p$), similar to the structure of CGAN~\cite{mirza2014conditional}. Using the poses from current video clips as input, the objective of pose generator ($G_p$) is to generate plausible future poses. The feature from the last convolutional layers of encoder in $G_p$ serve as the source of the dynamic motion feature $f_d$. The structure of the pose prediction network is illustrated in the upper half of Figure \ref{fig:pipline}.

Suppose $p_t\in \mathbb{R}^{2N}$ denotes the human pose captured in the $t$-th video frame, where $N$ is the number of skeleton joints. Given a randomly selected $T$-frame video clip, we apply OpenPose human skeleton detector \cite{cao2017realtime} to estimate the current pose sequence $P_{in}=(p_0,p_1, \cdots ,p_{T-1})\in \mathbb{R}^{1\times (T\times 2N)}$ and the future pose sequence $P_{gt}=(p_T,p_{T+1},...,p_{T+T'-1})$ of length $T'$ frames. 
The input to $G_p$ is a latent noise vector $z_p$ and the current pose sequence $P_{in}$. 
After applying a one-dimensional filter along the time direction, $G_p$ produces a sequence of poses for the next $T'$ frames, denoted as $P_{gen}=(p'_T,p'_{T+1},...,p'_{T+T'-1})$. 
During training, $G_p$ learns to generate pose sequences that match the target distribution, while $D_p$ aims to distinguish real pose sequences $P_{gt}$ from generated ones $P_{gen}$. The objective function is formulated as:

\begin{equation}
  \begin{split}
  \min_{G_p} &\max_{D_p} \mathcal{L}_{pose} = \\
  & \mathbb{E}_{p \sim P_{gt}}[\log D_p(P_{gt})]+ \mathbb{E}_{z \sim p(z)}[\log (1-D_p(G_p(z_p|P_{in}))]+\\
  & \frac{1}{T'}
  \sum^{T+T'-1}_{t=T}||p_t-p'_t||_2^2 + 
  \mathbb{E}_p[(\parallel \bigtriangledown D_p(P|P_{in})\parallel_2-1)^2]
  \end{split}
\label{pose_loss}
\end{equation}

The objective function mainly consists of two parts: L2 reconstruction loss contributes to the generation of accurate future poses, and the adversarial loss helps $G_p$ to learn the underlying distribution of the input data, leading to more realistic poses. 
In addition, we introduce the WGAN-GP gradient penalty \cite{gulrajani2017improved} (denoted as $E_p[(\parallel \bigtriangledown D_p(P|P_{in})\parallel_2-1)^2]$ in Equation \ref{pose_loss}) and OSGANs parameter update strategy \cite{shen2021osgan} to improve training stability and efficiency. Subsequently, we add convolutional layers and MLPs after the last layer in encoder of $G_p$ to keep the dimension of dynamic motion feature $f_d$ align with static context feature $f_s$. 

\subsection{Discriminative Context Matching Branch}

Besides the dynamic motion information, the context information is also significant for human action analysis. Thus, we conduct a discriminative context matching branch to learn static context features in self-supervised manner. 

As depicted in the blue box of Figure \ref{fig:pipline}, the V-Network (Video Network) and I-Network (I-frame network) are used to extract video and context features, respectively. We extract features $x_i \in \mathbb{R} ^{C_1 \times T_1\times H_1\times W_1}$ from a random video clip $i$ and feature $z_i \in \mathbb{R} ^{C_2\times H_2\times W_2}$ from the corresponding I-frame of the same video, where $C_1$, $T_1$, $H_1$, and $W_1$ denote the channel, time, height, and width of video clips, respectively. 
We use global average pooling to obtain their global representations $x_i^{pool}\in \mathbb{R} ^{C_1}$ and $z_i^{pool}\in \mathbb{R} ^{C_2}$ . We then apply two-layer MLP heads, $g^V$ and $g^I$, by taking the features of the video clip and I-frame as inputs, \ie $x_i^*=g^V(x_i^{pool})\in \mathbb{R} ^C$ and $z_i^*=g^I(z_i^{pool})\in \mathbb{R} ^C$. 
Finally, the Multi-Instance-InfoNCE (MI-InfoNCE) loss \cite{diba2021vi2clr} is applied to pull the video clip and I-frame feature pairs from the same video closer and push pairs from different video further away:
\begin{equation}
\label{MI-InfoNCE_loss}
  \begin{split}
  &\mathcal{L}_{context} =\\
  &-\frac{1}{B} \sum_{i=1}^B \log 
  \frac{\sum \exp (\frac{\cos(z^*_i,x^*_i)}{\tau})}
  {\sum \exp (\frac{\cos(z^*_i,x^*_i)}{\tau})+
  \sum_{k\neq i}^B \exp (\frac{\cos(z^*_k,x^*_i)}{\tau})}
  \end{split}
\end{equation}
where $z_i^*$ and $z_k^*$ denote the positive and negative I-frame features, $B$ denotes the number of samples in the minibatch, $\cos(z,x) = (z^T x)/(||z||_2\cdot||x||_2)$ denotes the cosine similarity between $z$ and $x$, and $\tau$ is a temperature adjusting the scale of cosine similarities. 
Considering that there may be multiple I-frames in one video clip, we set all I-frames in video clip $i$ as positive samples. Alternatively, the negative samples are I-frames from other videos in the same minibatch, denoted as $z^*_k$ where $k\neq i$.
Different from original InfoNCE loss \cite{infonce2018},  MI-InfoNCE loss enlarges the quantity of positive samples in each minibatch, which is more appropriate for the current situation. 
After pretraining on large video dataset, the video feature $x_i^*$ from the last layer of V-Network will serve as the static context feature $f_s\in \mathbb{R} ^C$ in our model. 

\subsection{Collaborative Video Generating Branch}
\label{sec:Videorecons}
After achieving the dynamic motion feature $f_d$ and tatic context feature $f_s$, we further conduct a collaborative video generating branch based on TGAN v2 \cite{saito2020tganv2} to learn more comprehensive video features by fusing these two features. 
This branch consists of three main components: a feature integration layer, a video reconstruction network and a video prediction network.

\begin{figure}[!hbt]
\centering
  \includegraphics[width=0.5\textwidth]{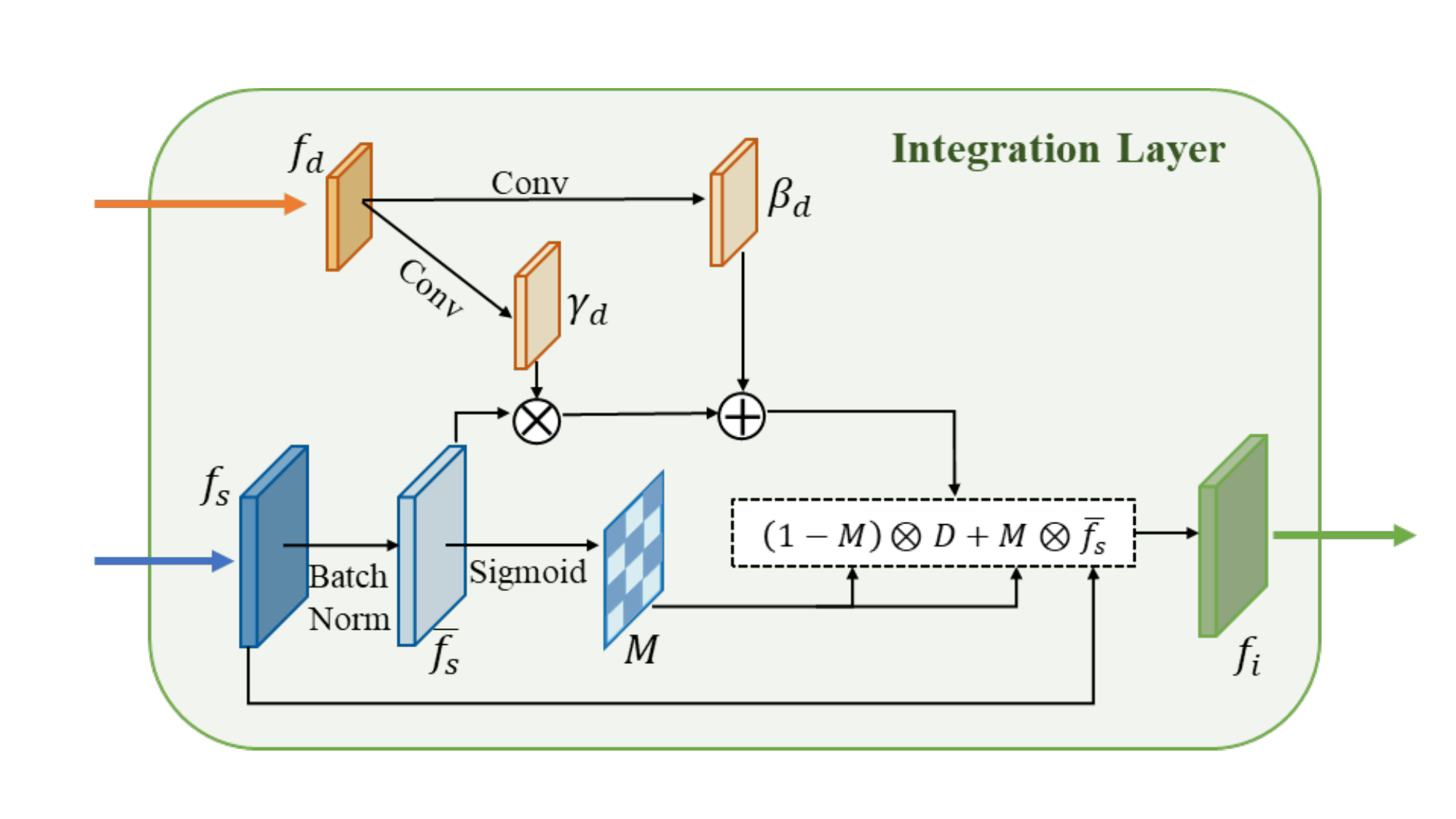}
  \caption{Detailed structure of integration layer. In the process of generating the integrated feature $f_i$, we use the parameters generated from $f_d$ to denormalize the normalized $\Bar{f_s}$, and combine the denormalized feature $D$ and static context feature $f_s$ with the an attention mask $M$.}
  \label{fig:adain}
\end{figure}

\noindent\textbf{Feature Integration Layer.}
Inspired by the prior work \cite{huang2017arbitrary, li2019faceshifter}, we adopt Adaptive Instance Normalization (AdaIN) for feature fusion. 
We firstly align the dimensions of the dynamic motion feature $f_d$ from the pose prediction branch and static context feature $f_s$ from V-network by adding convolutional layers and MLPs after the encoder of $G_p$. 
Then, the fusion of $f_d$ and $f_s$ is accomplished through a denormalization process. 
The detailed structure of the feature integration layer is shown in Fig. \ref{fig:adain}.

Specifically, we perform batch normalization \cite{ioffe2015batch} on the static context feature $f_s$:
\begin{equation}
  \begin{split}
  \Bar{f_s} = \frac{f_s-\mu}{\sigma}
  \end{split}
\end{equation}
where $\mu$ and $\sigma$ are the means and standard deviations of the channel-wise activations within an input minibatch. Then, we design two parallel branches from $\Bar{f_s}$ for dynamic motion integration and adaptive attention mask.

For dynamic motion integration, we compute the factor $D$ by denormalizing the normalized $\Bar{f_s}$, formulated as:
\begin{equation}
  \begin{split}
  D = \gamma_{d} \otimes \Bar{f_s} + \beta_{d}
  \end{split}
\end{equation}
where $\gamma_{d}$ and $\beta_{d}$ are modulation parameters obtained through convolution of the dynamic motion feature $f_d$. They share the same tensor dimensions with $\Bar{f_s}$.

The core issue of the integration layer is how to adaptively adjust the integration ratio of dynamic motion feature and static context feature to collaboratively facilitate network training. Therefore, we introduce an attention mechanism to the integration layer that dynamically adjusts the weights of $f_d$ and $f_s$ during the video reconstruction process. Specifically, we generate an attention mask $M$ from $\Bar{f_s}$ and normalize the mask via a Sigmoid function. 
Finally, the integrated feature $f_i$ is obtained by an element-wise combination of $D$ and $f_s$, weighted by the mask $M$, which is formulated as:
\begin{equation}
  \begin{split}
  f_i = (1-M)\otimes D + M \otimes f_s
  \end{split}
\end{equation}
Afterwards, the integrated feature $f_i$ is taken as the input of video generation networks. 
The following figure illustrates the detailed structure of integration layer.

\noindent\textbf{Video Generation Networks.}
We employ the video generating branch to learn dynamic motion feature and static context feature collaboratively. Our framework adopts TGAN v2, a time-efficient video generation network based on CGAN.

In the green box of Figure \ref{fig:pipline}, the video generating branch contains two similar TGANs, named video predictor and video reconstructor. 
The video predictor consists of a generator and a discriminator, denoted as $G^p_v$ and $D^p_v$, respectively. Taken the integrated feature $f_i$ and latent noise $z^p_v$ as input, the video predictor tries to generate future video clips. 
The video reconstructor has a similar structure, taking $f_i$ and latent noise $z^r_v$ as input, seeks to reconstruct current video clips simultaneously. As shown in the upper half of the green box in \ref{fig:pipline}, the generator and discriminator of video reconstructor are denoted as $G^r_v$ and $D^r_v$, respectively. 
We utilize adversarial loss and L2 loss for each TGAN, where the L2 reconstruction loss calculates the difference between the generated sample and the ground truth at the pixel level. The loss objective for video generating branch $\mathcal{L}_{video}$ involve two objectives $\mathcal{L}_{r}$ and $\mathcal{L}_{p}$: 

\begin{equation}
  \begin{split}
  \min_{G^r_v} &\max_{D^r_v} \mathcal{L}_{r} = \\
  & \mathbb{E}_{v\sim V_{r}}[\log D^r_v(V_{r})] + \mathbb{E}_{z\sim p(z)}[\log (1-D^r_V(G^r_V(z^r_v|f_i))]+\\
  & \frac{1}{THW}
  \sum^{THW}||v_{t}^r(h,w)-g_t^r(h,w)||_2^2
  \end{split}
\end{equation}

\begin{equation}
  \begin{split}
  \min_{G^p_v} &\max_{D^p_v} \mathcal{L}_{p} = \\
  & \mathbb{E}_{v \sim V_{p}}[\log D^p_v(V_{p})] + \mathbb{E}_{z \sim p(z)}[\log (1-D^p_V(G^p_V(z_v^p|f_i))]+\\
  & \frac{1}{THW}
  \sum^{THW}||v_{t}^p(h,w)-g_t^{p}(h,w)||_2^2
  \end{split}
\end{equation}

\begin{equation}
  \begin{split}
 \mathcal{L}_{video} = (\mathcal{L}_{r} + \mathcal{L}_{p})/2
   \end{split}
\end{equation}
where $V_{r}$ and $V_{p}$ denote the generation target of each generator videos and $T$, $H$, $W$ represent the temporal and spatial dimensions of video clip, respectively. $v_{t}^r(h,w)$ and $v_{t}^p(h,w)$ indicate the pixel value at the $(h,w)$ coordinates on the $t$-th frame in the ground truth video. The generated videos are denoted as $g_t^r(h,w)$ and $g_t^{p}(h,w)$. Similar to pose prediction branch, we use the OSGANs parameter updating strategy \cite{shen2021osgan} to simplify the training process. 
Two parallel training objectives enhance the understanding of the temporal relationship of the video, leading to a more comprehensive video representation.

\subsection{Joint Optimization}
In our CSVR, we jointly optimize the three branches by linearly combining the losses from each branch:
\begin{equation}
  \begin{split}
\mathcal{L}=\sigma_p\mathcal{L}_{pose}+\sigma_c\mathcal{L}_{context}+\mathcal{L}_{video}
  \end{split}
\end{equation}
where the  $\sigma_p, \sigma_c$ are scalar hyper-parameters. 
Since our CSVR contains three branches, we firstly pretrain each branch separately and then jointly optimize them with $\mathcal{L}$ in an end-to-end fashion for stable training. Finally, for fine-tuning on downstream tasks, we utilize the pretrained integrated feature $f_i$ as the initialization.

\section{Experiments}
In this section, we conduct experiments on four popular action recognition datasets, \ie UCF101 \cite{ucf101}, HMDB51 \cite{hmdb51}, Something-Something v2(SSv2) \cite{goyal2017ssv2} and Kinetics-400 \cite{kay2017kinetics} to verify the effectiveness of our CSVR. Besides, we also conduct ablation studies to investigate the significance of each branch in our method.

\subsection{Dataset}
The experiments of CSVR are mainly conducted on four datasets. UCF101 contains 13,320 videos covering 101 action classes, while HMDB51 consists of 6,766 videos in 51 classes. Following previous methods \cite{qian2021spatiotemporal, huang2021self}, we use split 1 of UCF101 and HMDB51 in our experiments. SSv2 is another extensive video dataset, emphasizes temporal information with its 169K training and 20K validation videos across 174 action classes. Kinetics-400 is a large-scale dataset containing 246K videos in its training set which includes 400 classes of human actions. 
During the experiments, we pretrain our CSVR on UCF101 training set and Kinetics-400 training set, respectively. Then we finetune the network initialized with the pretrained $f_i$ to two downstream tasks, \ie, action recognition and video retrieval, on UCF101 and HMDB51, respectively.

\subsection{Implementation Details}
\noindent\textbf{Backbone} 
For context matching branch, we explore four 3D-CNN backbones as V-Network, \ie, C3D \cite{c3d2015}, R(2+1)D\cite{resnet2p1d_2018}, R3D \cite{resnet2016,tran2018closer}, and S3D \cite{s3d2018} to extract video representations. Moreover, we adopt R2D-10 as I-Network to extract features from I-frame. For pose prediction branch, we utilize 1D-CNN-based CGAN for pose prediction. In addition, R3D-based TGAN v2 is leveraged in video generating branch. It should be noted that during inference, the major computation cost is from video representation extraction network which is mainly determined by various 3D-CNN backbones.

\noindent\textbf{Pretraining} 
Since our CSVR consists of three branches, we firstly pretrain each branch separately and then jointly optimize them in self-supervised manner for stable training. 
Firstly, CGAN in pose prediction branch and TGAN v2 in video generating branch are pretrained with respective training objectives in unsupervised manner. Then the whole CSVR network is pretrained in self-supervised manner, aiming for a video network with comprehensive feature. During self-supervised learning progress, integrated feature $f_i$ is utilized as a conditional variable to drive the generators in producing diverse videos. It is worth noting that GANs are not frozen during self-supervised pretraining.

For self-supervised pretraining, we randomly crop each input video clip to 16 frames with a temporal stride of 4 and resize them to 112$\times$112. We use a batch size $B$ of 16 and set the learning rate to $0.0005\times B$. We use SGD as the optimizer, setting the weight decay to 0.005 and the momentum to 0.9.
In each training step of context matching branch, we sample all the I-frames as positive samples and three negative frames from other video. Data augmentation approaches, \ie, random crop, random flip, Gaussian blur, and color jitter are conducted on video clips and I-frames for better performance.

As for the generation of $f_i$, the feature of the last layer in pose generator is utilized as the the dynamic motion feature $f_d$.
After investigating the selection of hyper parameters, which is introduced in Section \ref{sec:AblationStudy}, we set $\sigma_p=\sigma_c=0.75$ in our final experiment. 

\noindent\textbf{Finetuning}
After pretraining our CSVR on video dataset, the integrated feature $f_i$ is utilized to finetune on downstream tasks. We conduct experiments on two widely-used human action datasets, \ie, UCF101 and HMDB51, and report Top-1 accuracy of action recognition task and Recall at k (R@k) of video retrieval task on each dataset.

For action recognition task, we initialize the backbone with the pretrained parameters except for the last fully connected layer.
We finetune the model with a batch size $B$ of 32 and use the cosine annealing scheduler to decay the learning rate during training. 
For UCF101, we set the learning rate to $lr = 0.0001 \times B$ and the weight decay of the SGD optimizer to 0.003. For HMDB51, we set the learning rate to $lr = 0.0002 \times B$ and the weight decay to 0.002. 
Pretrain models are finetuned on both datasets for 100 epochs.
We incorporate a dropout ratio of 0.3 in the second-to-last layer of the network during finetuning.
Data augmentation methods, \eg, random crop, random flip, Gaussian blur, and color jitter are applied to increase the diversity of the training data. 

For the video retrieval task, we sample 10 video clips with a sliding window and use the average of their global features as the representation of the video.
The representations of videos from the test set are utilized to query the k-nearest neighbours in the training set, and Recall at k (R@k) is utilized for the evaluation metric. The network settings, \eg, batch size, learning rate and dropout ratio, are kept the same as the action recognition task. During retrieval, we directly use the representation from the pretrained video network for evaluation. For this task, we do not apply any augmentation except for random crop. 

\subsection{Ablation Study}
\label{sec:AblationStudy}
To further investigate the effectiveness of each branch of CSVR, we conduct experiments by firstly pretraining on UCF101 training set and then finetuning on the same set for action recognition. We choose R(2+1)D as the backbone of V-network and finetuning network for all ablation studies. 

\begin{table}[!t]
  \centering
    \caption{Action recognition performance comparison of different branches. In table below, $f_i$(AdaIN) and $f_i$(Concate) represents the integrated feature $f_i$ generated by AdaIN or directly concatenating, both of which are generated without optimization from video generating branch. $f_i$(full) indicates $f_i$ learned by the full CSVR pretraining framework.}
  \begin{tabular}{ccccc|cc}
    \bottomrule
    $f_d$ & $f_s$ & $f_i$(AdaIN) & $f_i$(Concat) & $f_i$(full) & UCF101 & HMDB51\\
    \hline
    \multicolumn{4}{l}{(Training from scratch)} && 65.0	& 32.5\\
    \checkmark &  &  &&  & 75.1& 38.4\\
     & \checkmark &  &&  & 73.1	& 40.2\\
    &  & \checkmark & &&  81.0	& 46.3\\
     &  &  & \checkmark&&  75.2	& 41.7\\
    &  &  & &\checkmark& \textbf{90.4}	& \textbf{56.6}\\
    \bottomrule
  \end{tabular}

  \label{table:parts_ablation}
\end{table}

\noindent\textbf{Effectiveness of Static and Dynamic Features.}
Since both static information and dynamic motion are crucial to understanding actions, we evaluate the contribution of features from the pose prediction branch and the context matching branch to the performance improvement in this experiment. Table \ref{table:parts_ablation} shows the final action recognition performance on UCF101 and HMDB51 with features from various branches as the initialization for finetuning on downstream action recognition. As seen, both the static context feature $f_s$ and dynamic motion feature $f_d$ significantly improve the action recognition accuracy compared to the baseline of training from scratch, demonstrating the superiority of designing context matching and pose prediction branches. Interestingly, the dynamic motion feature $f_d$ achieves better results than static context feature $f_s$ on HMDB51, while $f_d$ underperform $f_s$ on UCF101. 
This may be because most classes in UCF101 can be classified by context information, while classes in HMDB51 are mainly distinguished by the motion information. 

\noindent\textbf{Effectiveness of Video Generating Branch.}
The collaborative video generating branch contains two components: the integration layer and the video generating networks. 
Following previous practice, we evaluate the effectiveness of these components by finetuning the feature on action recognition tasks on UCF101 and HMDB51.
By default, the integrated feature $f_i$ is generated by fusing the static context feature $f_s$ and dynamic motion feature $f_d$ with AdaIN. 
As seen in Table. \ref{table:parts_ablation}, when employing $f_i$ for the downstream task finetuning, the improvement on accuracy further illustrates the complementarity of static context and dynamic motion features for action recognition. 
To further emphasize the effectiveness of AdaIN, we directly concatenate \( f_d \) and \( f_s \), followed by convolution to generate the concat feature $f_i$(Concat). It can be observed that the concat feature offers slight improvements over those produced by previous branches in downstream tasks and performs much worse than the feature generated by AdaIN.
Finally, we use the full CSVR framework to further improve the integrated feature $f_i$ by jointly optimizing three branches. The accuracy on UCF101 and HMDB51 is significantly improved to 90.3\% and 56.5\%, respectively, which shows the effectiveness of collaboratively learning static context and dynamic motion features for action recognition.
\setlength{\tabcolsep}{5pt}
\begin{table}[!t]
\centering
\caption{
   Correlation between pretext and downstream tasks. 
   The linear relationship between the results indicates the effectiveness of our pretraining.
   MSE score is increased by a factor of 100 to emphasize the change. All the experiments are conducted on UCF101 dataset.
}
\begin{tabular}{l|cccccc} \bottomrule
\multirow{2}{*}{Task (Evaluation)} & \multicolumn{6}{c}{Pretraining epoch} \\ \cmidrule{2-7}
& 1 & 5 & 10 & 20 & 50 & 120 \\ \hline
Context Matching (Acc$\uparrow$) & 36.3 & 47.0 & 49.9 & 52.9 & 52.0 & 63.9 \\
Pose Prediction (MSE$\downarrow$) & 32 & 16 & 5 & 4.2 & 3.4 & 2.9\\ 
Video Reconstruction (IS$\uparrow$) & 1.4 & 8.2 & 9.3 & 11.9 & 14.6 & 15.1 \\
\hline
Action Recognition(Acc$\uparrow$) & 65.0 & 72 & 82.4 & 83.5 & 88.3 & 89.6 \\ \bottomrule
\end{tabular}

  \label{table:prgress_ablation}
\end{table}

\noindent\textbf{Correlation between pretext and downstream tasks.}
We conduct experiments to explore the correlation between the pretext tasks and the downstream task by measuring their performance during different pretraining epochs. The performance measurements include the top-1 accuracy of context matching, mean squared error (MSE) of pose prediction, inception score (IS) of generated videos and top-1 accuracy of downstream action recognition. The results in Table \ref{table:prgress_ablation} show a consistent improvement in action recognition accuracy as the performance of the pretext tasks increases, which indicates the strong correlations between various pretext and downstream tasks.

\begin{table}[h]
\begin{minipage}[b]{0.45\linewidth} 
\centering
\caption{Comparison of feature $f_d$ from different layers of $G_p$ encoder.}
\resizebox{\linewidth}{!}{ 
    \begin{tabular}{c|cc} 
    \hline
    Layer & UCF101 &  HMDB51 \\ 
    \hline
    Layer 2	& 72.0 & 36.9 \\
    Layer 4	& 74.5 & 37.2 \\
    Layer 6	& 75.1 & 38.4	\\
    \hline 
    \end{tabular}
    }
\label{reb:Gp_layers}
\end{minipage}
\hfill
\begin{minipage}[b]{0.5\linewidth}
\centering
\caption{Comparison of feature $f_s$ with different loss functions.}
\resizebox{\linewidth}{!}{ 
    \begin{tabular}{c|cc} 
    \hline
    Objective & UCF101 &  HMDB51 \\ 
    \hline
    L2 Loss	& 70.2 & 33.4	\\
        InfoNCE	& 72.5 & 40.2 \\
    MI-InfoNCE	& 73.1 & 40.2 \\
    \hline 
    \end{tabular}
    }
\label{reb:InfoNce}
\end{minipage}
\end{table}

\noindent\textbf{Selection of Dynamic Motion Feature.} 
In order to locate the hidden layer which enriches the most dynamic information, we pretrain the pose generator $G_p$ on poses from UCF101 dataset and use features from different layers of encoder in $G_p$ as dynamic motion features $f_d$.
We report the results of different $f_d$ finetuned on action recognition task. As shown in Table \ref{reb:Gp_layers}, $f_d$ from the last layer achieves the best performance. 

\noindent\textbf{Selection of Contrastive Loss.}
As shown in the lower half of Figure \ref{fig:pipline}, we employ a contrastive loss between video feature and I-frame feature. 
The widely used InfoNCE loss \cite{infonce2018} compares the feature with limited positive samples and multiple negative samples, which can be formalized as: 
\begin{equation}
  \begin{split}
  &\mathcal{L}_{context} =\\
 & -\frac{1}{B} \sum_{i=1}^B \log 
  \frac{\exp (\frac{\cos(z^*_i,x_i)}{\tau})}{\exp (\frac{\cos(z^*_i,x_i)}{\tau})+\sum_{k \neq i}^B \exp (\frac{\cos(z^*_k,x_i)}{\tau})}
  \end{split}
\end{equation}
where $x_i$ denotes the global video representation, and $z_i^*$ and $z_k^*$ denote the positive and negative I-frame features, other symbols are the same as Equation \ref{MI-InfoNCE_loss}. However, there may exist multiple I-frames in a single video clip, indicating more positive samples.
Hence, we apply more advanced Multi-Instance-InfoNCE (MI-InfoNCE) \cite{diba2021vi2clr} as the contrast loss in static context branch. Comparing to the traditional InfoNCE loss utilized in \cite{huang2021self}, MI-InfoNCE loss enables more adequate comparison between positive and negative samples, resulting in better performance on downstream tasks. 
To further verify the effectiveness of contrastive loss, we also conduct ablation experiments with L2 loss.
When calculating InfoNCE and L2 loss, we randomly select an I-frame from the video clip as learning objective to ensure that the loss function produces correct outputs.
As shown in Table \ref{reb:InfoNce}, MI-InfoNCE achieve optimal results among three losses. Interestingly, we observe that both InfoNCE and MI-InfoNCE achieve identical scores when testing on HMDB51, while MI-InfoNCE outperforms the others on UCF101. This is due to action recognition on UCF101 relies more on static context information.

\noindent\textbf{Selection of Video Generation Target}
Since our video generating branch is trained with two different objective functions, we present the performance of each separately on the downstream tasks. Therefore, we train models in three different setups, \ie, jointly train CSVR with both $\mathcal{L}_{r}$ and $\mathcal{L}_{p}$, and train CSVR with $\mathcal{L}_{r}$ or $\mathcal{L}_{p}$ independently. As shown in Table \ref{reb:reconstruction objective}, the results obtained from the full branch objective surpass the result of reconstruction objective in both cases. Moreover, when compared to the result of prediction objective, the full branch objective outperforms better on the UCF101 dataset and performs equally well on HMDB51. When conducting experiments on the SSv2 dataset, which is more sensitive to temporal information, the performance using the full branch objective significantly surpassed that of using $\mathcal{L}_{r}$ or $\mathcal{L}_{p}$ independently.

\begin{table}[hbt!]
\caption{Performance comparison of using different objective functions in video generating branch to train CSVR.}
\centering
    \begin{tabular}{l|ccc} 
    \hline
    Objective & UCF101 &  HMDB51 & SSv2 \\ 
    \hline
    $\mathcal{L}_{r}$ 	& 89.2 & 54.6& 56.1	\\
    $\mathcal{L}_{p}$	& 89.9 & 56.6 & 55.3\\
    $\mathcal{L}_{r}$ \& $\mathcal{L}_{p}$	& 90.4 & 56.6& 59.2 \\
    \hline 
    \end{tabular}
\label{reb:reconstruction objective}
\end{table}

\begin{figure}[!hbt]
\centering
  \includegraphics[width=0.8\linewidth]{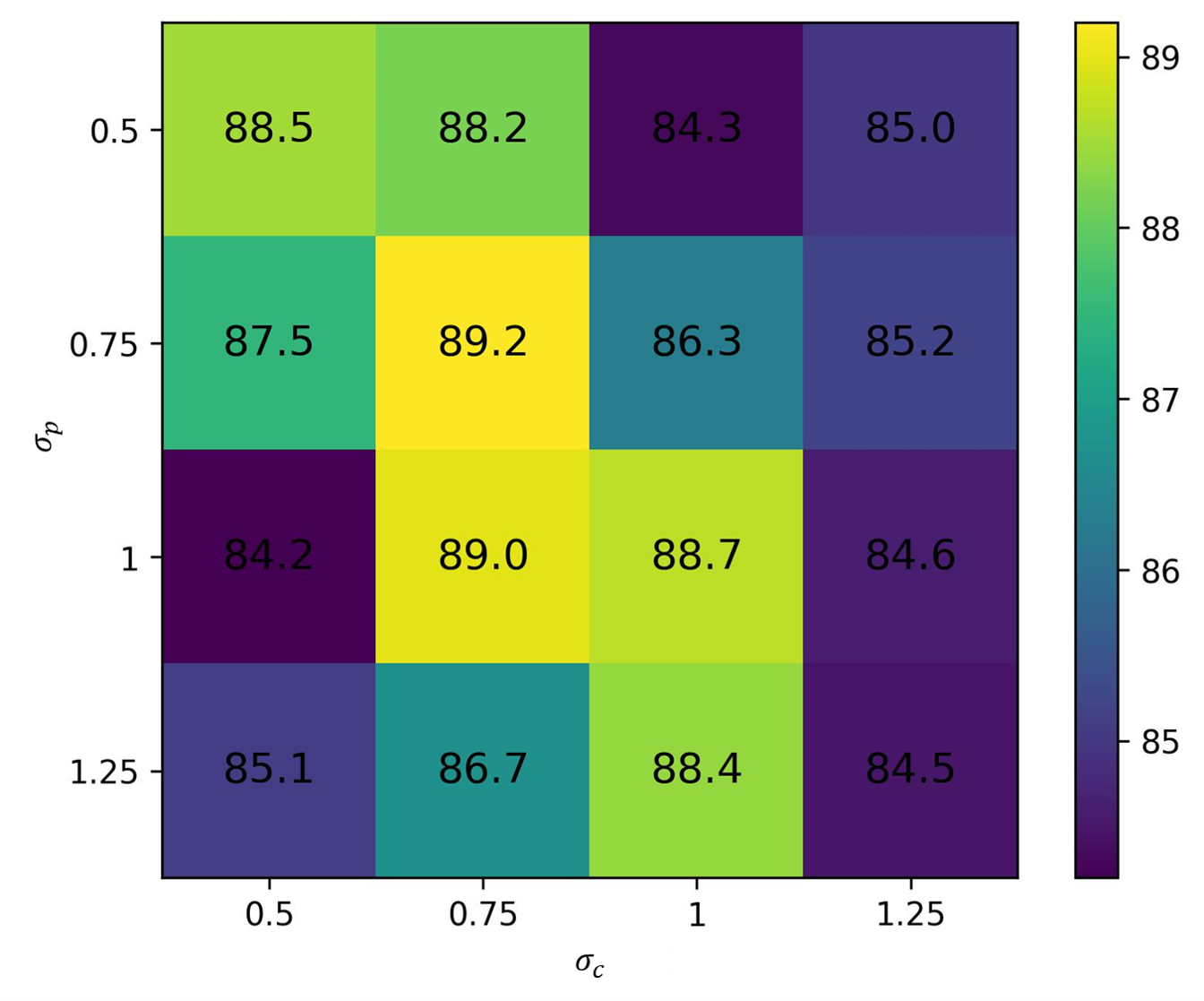}
  \caption{Action recognition results on UCF101 with different combinations of $\sigma_p,\sigma_c$.}
  \label{fig:heatmap}
\end{figure}

\noindent\textbf{Hyperparameter Investigation of $\sigma_p, \sigma_c$.}
CSVR contains three branches which are jointly optimized by loss function $L$. $\sigma_p, \sigma_c$ in $L$ are the hyperparameters that coordinate the weights of pose prediction branch and context matching branch, respectively. 
We perform a grid search on all combinations of two parameters within a range of 0.5 - 1.25 and conduct the pretraining experiments with each combination to find the optimal parameters. 
Figure \ref{fig:heatmap} shows the results on UCF101 when transferring the pretrained features to the downstream action recognition task.  
As seen, the best results are achieved when $\sigma_p = \sigma_c = 0.75$, which are chosen for all other experiments.

\begin{figure*}[!htb]
  \centering
  \includegraphics[width=0.8\textwidth]{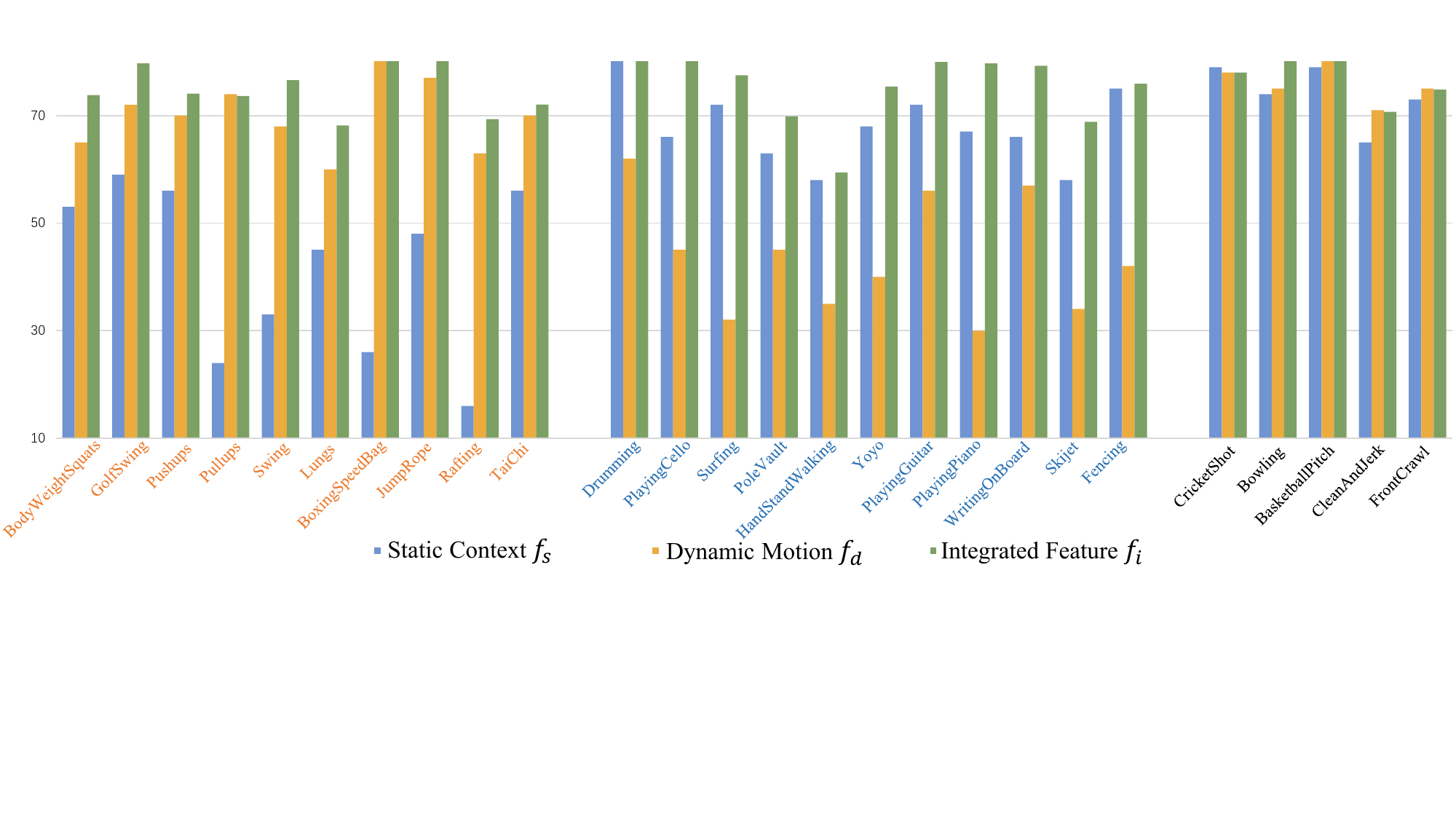}
  \vspace{-2.5cm}
  \caption{Per-class action recognition accuracy with different features. Labels on the horizontal axis are divided into three groups: the orange group are highly related to motion information, the blue group can be easily recognized by static appearance, the black group exhibit both fixed action patterns and rich scene changes. The figure reveals the contributions of different features in action recognition.}
  \label{fig:per-big}
\end{figure*}

\noindent\textbf{Per-Class Accuracy of Action Recognition with Different Features.}
To further investigate the contribution of features from different branches to action understanding, we present a per-class accuracy visualization of typical 26 classes from UCF101 in Figure \ref{fig:per-big}. The action classes are divided into three groups shown on horizontal axis. The group represented by orange characters (denoted as "orange group") are highly related to dynamic motion. The group represented by blue characters (denoted as "blue group") can be easily recognized by static context. And the group represented by black characters (denoted as "black group") need both motion and context to distinguish. For the orange group, the dynamic motion feature $f_d$ achieves much better results than the static context feature $f_s$, while the opposite conclusion is obtained for the blue group. This result demonstrates that the carefully designed pretext tasks can well capture different types of features beneficial to action recognition task. 
Moreover, by jointly optimizing three branches, the integrated feature $f_i$ is refined and greatly improves the feature representation for actions, whether based on motion or context.

\subsection{Comparison with State-of-the-art Methods}
To verify the effectivenss of our CSVR framework, we conduct experiments on two downstream tasks, \ie, action recognition and video retrieval, with four 3D-CNN backbones, \ie, R(2+1)D, R3D, C3D and S3D. The whole model is pretrained on two datasets, \ie, UCF101 and large-scale Kinetics-400, respectively. 
\setlength{\tabcolsep}{8pt}
\begin{table*}[!htb]
\small
\centering
\caption{\textbf{Action Recognition on UCF101 and HMDB51}. Comparison of CSVR with state-of-the-art self-supervised methods on the action recognition task. *STCNet is a specially-designed R3D network used in DynamoNet\cite{diba2019dynamonet}.}
\begin{tabu}{lc|cccc|cc} \hline
Method & Backbone & Pretrain Dataset & Epochs & Resolution & Clip Length & UCF101 & HMDB51 \\ 
\hline
DSM\cite{dsm2020} & C3D & UCF101 & 200 & $112\times 112$ & 16 & 70.3 & 40.5 \\
CoCLR\cite{han2020self} & S3D	& UCF101	& 700 & $128\times 128$	& 32	& 81.4	& 52.1\\
MSCL\cite{ni2022motion}	& S3D	& UCF101	& 200 & $128\times 128$	& 32	& 83.7	& 53.8\\
DCM\cite{huang2021self}	& R(2+1)D	& UCF101	& 120 & $112\times 112$	& 16	& 79.7	& 48.6\\
VCL\cite{qian2022static}	& R(2+1)D	& UCF101	& 200 & $112\times 112$	& 16	& 82.1	& 49.7\\
TCLR\cite{dave2022tclr}	& R3D	& UCF101	& 400 & $112\times 112$	& 16	& 82.4	& 52.9\\
MSCL\cite{ni2022motion}	& R3D	& UCF101	& 600 & $112\times 112$	& 16	& 86.7	& 58.9\\
Vi$^2$CLR\cite{diba2021vi2clr}	& S3D	& UCF101 & 300 & $112\times 112$	& 16& 82.8 & 52.9\\
\hline
CSVR(Ours) & C3D & UCF101 & 200 & $112\times 112$ & 16 & 88.3 & 52.3 \\
CSVR(Ours) & S3D & UCF101 & 200 & $112\times 112$ & 16 & 88.1 & 54.9 \\
CSVR(Ours) & R(2+1)D & UCF101 & 200 & $112\times 112$ & 16 & \textbf{90.4} & 56.6  \\
CSVR(Ours) & R3D & UCF101 & 200 & $112\times 112$ & 16 & 89.6 & \textbf{60.7}\\
\hline
\hline
DCM\cite{huang2021self} & C3D & Kinetics-400 & 120 & $112\times112$ & 16 & 83.4 & 52.9\\
CoCLR\cite{han2020self}	& S3D	& Kinetics-400	& 500 & $128\times 128$	& 32	& 87.9	& 54.6\\
VideoMoCo\cite{pan2021videomoco}	& R(2+1)D	& Kinetics-400	& 200 & $112\times 112$	& 32	& 78.7	& 49.2\\
DCM\cite{huang2021self}	& R(2+1)D	& Kinetics-400	& 120 & $112\times 112$	& 16	& 85.7	& 54.0\\
FAME\cite{ding2022motion}	& R(2+1)D	& Kinetics-400 & 200 & $112\times 112$	& 16	& 84.8	& 53.5\\
VCL\cite{qian2022static}	& R(2+1)D	& Kinetics-400	&100 & $112\times 112$	& 16	& 86.1	& 54.8\\
TCLR\cite{dave2022tclr}	& R3D	& Kinetics-400	& 100 & $112\times 112$	& 16	& 84.1	& 53.6\\
TE\cite{jenni2021time}	& R3D	& Kinetics-400	& 200 & $128\times 128$	& 16	& 87.1	& 63.6\\
DynamoNet\cite{diba2019dynamonet}	& STCNet*	& Kinetics-400 & 200 & $112\times 112$	& 16& 88.1	& 59.9\\
Vi$^2$CLR\cite{diba2021vi2clr}	& S3D	& Kinetics-400 & 300 & $112\times 112$	& 16& 89.1 & 55.7\\
MSCL\cite{ni2022motion}	& R3D	&Kinetics-400	& 400 & $112\times 112$	& 16	& 90.7	& 62.3\\
KnowledgeGuide\cite{wang2024knowledge}	& C3D & Kinetics-400	& 400 & $112\times 112$	& 16	& 82.7	& 47.0\\
KnowledgeGuide\cite{wang2024knowledge}	& R(2+1)D & Kinetics-400	& 400 & $112\times 112$	& 16	& 82.4	& 46.7\\
\hline
CSVR(Ours) & C3D & Kinetics-400 & 200 & $112\times 112$ & 16 & 92.8 & 62.8 \\
CSVR(Ours) & S3D & Kinetics-400 & 200 & $112\times 112$ & 16 & 90.2 & 61.2 \\
CSVR(Ours) & R(2+1)D &   Kinetics-400 & 200 & $112\times 112$ & 16 & 92.8 & 64.0 \\
CSVR(Ours) & R3D &  Kinetics-400 & 200 & $112\times 112$ & 16 & \textbf{94.5} & \textbf{65.5} \\
\hline 
\end{tabu}
\label{tab:sota_recognition}
\end{table*}

\noindent\textbf{Action Recognition on UCF101 and HMDB51.}
We firstly compare CSVR with the previous self-supervised video representation learning methods in terms of the downstream action recognition task. 
As shown in Table \ref{tab:sota_recognition}, when pretraining on UCF101, our CSVR consistently outperforms all existing methods with the same backbone. Moreover, when pretraining on large-scale dataset Kinetics-400, our method also achieves the best results of $94.5\%$ and $65.5\%$ on UCF101 and HMDB51, respectively. 
We can also observe that the trained model by our CSVR surpasses the methods which use other modality of data like optical flow \cite{han2020self}, multi-modal information \cite{huang2021self}.
Besides, our CSVR needs fewer pretraining epochs than most existing methods.
All these results demostrate that CSVR is both effective and efficient for self-supervised video representation learning on action recognition.


\setlength{\tabcolsep}{8pt}
\begin{table*}[!htb]
\small
\centering
\caption{\textbf{Action Recognition on SSv2}. Comparison of CSVR with state-of-the-art self-supervised methods on the action recognition task. }
\begin{tabu}{l|ccccc|cc} \hline
Method & Backbone & Pretrain Dataset & Epochs & Resolution & Clip Length & Top-1 \\ 
\hline
VideoMoCo\cite{pan2021videomoco}	& R(2+1)D	& Kinetics-400	& 200 & $112\times 112$	& 32 & 19.5	\\
$\rho$MoCo\cite{feichtenhofer2021large}	& SlowFast-R50	& Kinetics-400 & 200 & $224\times224$	& 16 & 54.4	\\
CORP\cite{hu2021contrast}& I3D	& Kinetics-400 & 800 & N/A	& 16 & 48.8\\
SVT\cite{ranasinghe2022self}	& ViT-B	& Kinetics-400 & 20 & $224\times224$	& 16 & \textbf{59.2}	\\
\hline
CSVR(Ours) & R(2+1)D &   Kinetics-400 & 200 & $112\times 112$ & 16 & \textbf{59.2} \\
\hline 
\end{tabu}
\label{tab:sota_recognition_ssv2}
\end{table*}

\noindent\textbf{Action Recognition on SSv2.}
Comparing to UCF101 and HMDB51, SSv2 exhibits less scene bias than previous datasets and require a stronger temporal understanding to accurately classify actions. As shown in Table. \ref{tab:sota_recognition_ssv2}, our method outperforms the best previous methods using 3D-CNN backbones by an absolute margin of 4.8\% on SSv2 when pretrained on Kinetics-400. 
In addition, the performance of CSVR on the SSv2 action recognition task is comparable to that of SVT\cite{ranasinghe2022self}, which uses a stronger ViT backbone\cite{dosovitskiy2020image} and higher video resolution. 
Such results demonstrating its suitability for video datasets captured in controlled, real-world settings.

\begin{table*}[!tb]
\small
\centering
\caption{\textbf{Video Retrieval.} Comparison of CSVR with the state-of-the-art on nearest-neighbour video retrieval on UCF101 and HMDB51. Given query test clips, our goal is to find training clips that are from the same class using Recall at $k$(R@$k$) metric.}
\begin{tabular}{lcc|cccc|cccc}
\hline
\multirow{2}{*}{Method} & \multirow{2}{*}{Backbone} & \multirow{2}{*}{Epochs} & \multicolumn{4}{c|}{UCF101} & \multicolumn{4}{c}{HMDB51} \\ 
\cmidrule{4-11}
& & & R@1 & R@5 & R@10 & R@20  & R@1 & R@5 & R@10 & R@20 \\ 
\hline
TCGL\cite{liu2022tcgl}	& R(2+1)D	& 300	& 21.5	& 39.3	& 49.3	& 59.5	& 10.5	& 27.6 & 39.7 & 55.6\\
MCN\cite{lin2021self}	& R(2+1)D	& 500 & 52.5	& 69.5	& 77.9	& 83.1	& 23.7 & 46.5	& 58.9	& 72.4\\
MotionFit\cite{gavrilyuk2021motion}	& R(2+1)D	& 120	& 61.6	& 75.6	& -	& 85.5	& 29.4	& 46.5 & - & 66.7\\
DualVar\cite{zhang2021inter}	& R3D & 200	& 46.7	& 63.1	& 69.7	& 78.0	& -	& -	& -	& -\\
MSCL\cite{ni2022motion}	& R3D & 400	& 63.7	& 79.1	& 84.0	& -	& 32.6	& 58.5	& 70.5	& -\\
KnowledgeGuide\cite{wang2024knowledge}& R(2+1)D & 400	& 52.6	& 70.1	& 77.4	& 85.2	& 22.2	& 45.9	& 59.9	& 74.2\\
KnowledgeGuide\cite{wang2024knowledge}& C3D & 400	& 49.9	& 68.3	& 77.4	& 84.9	& 22.4	& 46.9	& 60.3	& 73.1\\
\hline
CSVR(Ours) & R(2+1)D & 200 & \textbf{64.4} & \textbf{80.2} & 87.8 & \textbf{93.1} & 34.1 & 57.2 & 71.0 & 81.4\\
CSVR(Ours) & R3D & 200 & 64.2 & 79.5 & \textbf{88.2} & 92.9 & \textbf{34.5} & \textbf{59.0} & \textbf{71.8} & \textbf{82.2} \\
\hline
\end{tabular}
\label{tab:sota_retrieval}
\end{table*}

\noindent\textbf{Video Retrieval.}
Besides action recognition, we also conduct evaluations for the downstream video retrieval task on two human action datasets UCF101 and HMDB51. Table \ref{tab:sota_retrieval} shows the results of video retrieval of each method in terms of R@K, where R@K denotes the recall rate of top-K.
All methods are pretrained on Kinetics-400 with the resolution of 112$\times$112 for fair comparisons.
The similar conclusion can be obtained that our CSVR achieves the best results on both UCF101 and HMDB51, showcasing its superiority for self-supervised features learning for human action understanding.

\begin{table}[!thb]
\centering
\caption{Computing cost comparing with popular self-supervised video representation learning methods. All methods are pretrained on K400.}
\resizebox{\linewidth}{!}{ 
\begin{tabu}{l|ccc|c} 
\hline
Method & Backbone & GFLOPs&Parameters & UCF Top-1 \\ 
\hline
MoCo\cite{moco2019} & C3D & 38.7 &33.4M & 60.5\\
CoCLR\cite{han2020self} & S3D &36.0& 9.1M & 87.9\\
SpeedNet\cite{speednet2020} & S3D &36.5& 9.1M & 81.1\\
$\rho$MoCo\cite{feichtenhofer2021large} & SlowFast-R50 &83.5& 31.8M & 92.8\\
\hline
\textbf{CSVR} & C3D &40.4 & 33.4M & 92.8  \\
\textbf{CSVR} & S3D  &37.7& 9.1M & 90.2 \\
\textbf{CSVR} & R(2+1)D  &40.4& 15.4M & 92.8  \\
\hline
\end{tabu}
}
\label{tab:GFLOPs_sota}
\end{table}

\noindent\textbf{Computational cost.}
To compare computational costs, we enumerate the model parameters of some popular self-supervised video representation learning methods pretrained on Kinetics-400 in Table. \ref{tab:GFLOPs_sota}. Following the common practices in \cite{feichtenhofer2021large,lin2024tsgan,tong2022videomae}, we report the parameters of each model during the inference phase.
As shown, CSVR significantly improves accuracy in terms of the UCF101 action recognition downstream task compared to methods using the same backbone.
Compared to MoCo\cite{moco2019}, which also uses C3D as the backbone, our method boosts performance by over 30\% on the UCF101 dataset. Additionally, CSVR using R(2+1)D as the backbone achieves comparable performance with $\rho$MoCo\cite{feichtenhofer2021large}, which uses the heavier SlowFast \cite{feichtenhofer2019slowfast} as the backbone. 
Noting that, CSVR has higher GFLOPs compared to the methods using the same backbone since we have an additional generative pose prediction branch.

\section{Limitation}
CSVR is a self-supervised video representation learning framework with multiple branches. These branches independently learn dynamic and static features, which are then optimized collaboratively. To manage the complexity of optimizing multiple neural networks during pretraining, we opt for lightweight CNNs instead of stronger, but larger backbones like Transformers. Additionally, we see potential for integrating more advanced generative methods (\eg, diffusion models\cite{ho2020diffusion}) and contrastive learning approaches (\eg, BYOL\cite{feichtenhofer2021large}) into our framework, given that the primary goal of the dynamic and static branches is to learn comprehensive video representations. Exploring these enhancements further will be a focus of our future work.

\section{Conclusion}
This paper designs a self-supervised video representation learning framework named CSVR for action recognition by collaboratively training three branches: generative pose prediction branch, discriminative context matching branch and video generating branch. The pose prediction branch is responsible for extracting dynamic motion features, while the context matching branch focuses on learning static context features. Furthermore, the video generating branch jointly optimizes the dynamic motion and static context features, leading to a more comprehensive feature representation for the downstream tasks. Extensive experiments demonstrate that our CSVR outperforms the state-of-the-art methods on both action recognition and video retrieval tasks.
In the future, we will introduce more powerful backbone like Transformer into CSVR to further explore potential of the learning framework for various video understanding tasks. We hope our CSVR can inspire the researches on the collaboration of discriminative and generative self-supervised learning.


\bibliography{sn-bibliography}

\begin{thebibliography}{100}
\providecommand{\url}[1]{#1}
\csname url@samestyle\endcsname
\providecommand{\newblock}{\relax}
\providecommand{\bibinfo}[2]{#2}
\providecommand{\BIBentrySTDinterwordspacing}{\spaceskip=0pt\relax}
\providecommand{\BIBentryALTinterwordstretchfactor}{4}
\providecommand{\BIBentryALTinterwordspacing}{\spaceskip=\fontdimen2\font plus
\BIBentryALTinterwordstretchfactor\fontdimen3\font minus \fontdimen4\font\relax}
\providecommand{\BIBforeignlanguage}[2]{{%
\expandafter\ifx\csname l@#1\endcsname\relax
\typeout{** WARNING: IEEEtran.bst: No hyphenation pattern has been}%
\typeout{** loaded for the language `#1'. Using the pattern for}%
\typeout{** the default language instead.}%
\else
\language=\csname l@#1\endcsname
\fi
#2}}
\providecommand{\BIBdecl}{\relax}
\BIBdecl

\bibitem{ni2022motion}
J.~Ni, N.~Zhou, J.~Qin, Q.~Wu, J.~Liu, B.~Li, and D.~Huang, ``Motion sensitive contrastive learning for self-supervised video representation,'' in \emph{ECCV}.\hskip 1em plus 0.5em minus 0.4em\relax Springer, 2022.

\bibitem{gavrilyuk2021motion}
K.~Gavrilyuk, M.~Jain, I.~Karmanov, and C.~G. Snoek, ``Motion-augmented self-training for video recognition at smaller scale,'' in \emph{ICCV}, 2021.

\bibitem{qian2022static}
R.~Qian, S.~Ding, X.~Liu, and D.~Lin, ``Static and dynamic concepts for self-supervised video representation learning,'' in \emph{ECCV 2022}.\hskip 1em plus 0.5em minus 0.4em\relax Springer, 2022.

\bibitem{arnab2021vivit}
A.~Arnab, M.~Dehghani, G.~Heigold, C.~Sun, M.~Lu{\v{c}}i{\'c}, and C.~Schmid, ``Vivit: A video vision transformer,'' in \emph{ICCV}, 2021.

\bibitem{wu2018compressed}
C.-Y. Wu, M.~Zaheer, H.~Hu, R.~Manmatha, A.~J. Smola, and P.~Kr{\"a}henb{\"u}hl, ``Compressed video action recognition,'' in \emph{CVPR}, 2018.

\bibitem{feichtenhofer2022masked}
C.~Feichtenhofer, H.~Fan, Y.~Li, and K.~He, ``Masked autoencoders as spatiotemporal learners,'' in \emph{NeurIPS}, 2022.

\bibitem{zhang2021inter}
L.~Zhang, Q.~She, Z.~Shen, and C.~Wang, ``Inter-intra variant dual representations forself-supervised video recognition,'' \emph{arXiv preprint arXiv:2107.01194}, 2021.

\bibitem{martinez2021pose}
A.~Mart{\'\i}nez-Gonz{\'a}lez, M.~Villamizar, and J.-M. Odobez, ``Pose transformers (potr): Human motion prediction with non-autoregressive transformers,'' in \emph{ICCV}, 2021.

\bibitem{wang2019self}
J.~Wang, J.~Jiao, L.~Bao, S.~He, Y.~Liu, and W.~Liu, ``Self-supervised spatio-temporal representation learning for videos by predicting motion and appearance statistics,'' in \emph{CVPR}, 2019.

\bibitem{gui2023survey}
J.~Gui, T.~Chen, Q.~Cao, Z.~Sun, H.~Luo, and D.~Tao, ``A survey of self-supervised learning from multiple perspectives: Algorithms, theory, applications and future trends,'' \emph{arXiv preprint arXiv:2301.05712}, 2023.

\bibitem{huang2021ascnet}
D.~Huang, W.~Wu, W.~Hu, X.~Liu, D.~He, Z.~Wu, X.~Wu, M.~Tan, and E.~Ding, ``Ascnet: Self-supervised video representation learning with appearance-speed consistency,'' in \emph{ICCV}, 2021.

\bibitem{benaim2020speednet}
S.~Benaim, A.~Ephrat, O.~Lang, I.~Mosseri, W.~T. Freeman, M.~Rubinstein, M.~Irani, and T.~Dekel, ``Speednet: Learning the speediness in videos,'' in \emph{CVPR}, 2020.

\bibitem{wang2022bevt}
R.~Wang, D.~Chen, Z.~Wu, Y.~Chen, X.~Dai, M.~Liu, Y.-G. Jiang, L.~Zhou, and L.~Yuan, ``Bevt: Bert pretraining of video transformers,'' in \emph{CVPR}, 2022.

\bibitem{pan2021videomoco}
T.~Pan, Y.~Song, T.~Yang, W.~Jiang, and W.~Liu, ``Videomoco: Contrastive video representation learning with temporally adversarial examples,'' in \emph{CVPR}, 2021.

\bibitem{chen2021rspnet}
P.~Chen, D.~Huang, D.~He, X.~Long, R.~Zeng, S.~Wen, M.~Tan, and C.~Gan, ``Rspnet: Relative speed perception for unsupervised video representation learning,'' in \emph{AAAI}, 2021.

\bibitem{liu2022tcgl}
Y.~Liu, K.~Wang, L.~Liu, H.~Lan, and L.~Lin, ``Tcgl: Temporal contrastive graph for self-supervised video representation learning,'' \emph{IEEE TIP}, 2022.

\bibitem{song2018collaborative}
G.~Song and W.~Chai, ``Collaborative learning for deep neural networks,'' in \emph{NeurIPS}, 2018.

\bibitem{jenni2021time}
S.~Jenni and H.~Jin, ``Time-equivariant contrastive video representation learning,'' in \emph{ICCV}, 2021.

\bibitem{schneider2022pose}
D.~Schneider, S.~Sarfraz, A.~Roitberg, and R.~Stiefelhagen, ``Pose-based contrastive learning for domain agnostic activity representations,'' in \emph{CVPR}, 2022.

\bibitem{rai2021cocon}
N.~Rai, E.~Adeli, K.-H. Lee, A.~Gaidon, and J.~C. Niebles, ``Cocon: Cooperative-contrastive learning,'' in \emph{CVPR}, 2021.

\bibitem{shuffle_learn2016}
I.~Misra, C.~L. Zitnick, and M.~Hebert, ``Shuffle and learn: unsupervised learning using temporal order verification,'' in \emph{ECCV}.\hskip 1em plus 0.5em minus 0.4em\relax Springer, 2016.

\bibitem{opn2017}
H.-Y. Lee, J.-B. Huang, M.~Singh, and M.-H. Yang, ``Unsupervised representation learning by sorting sequences,'' in \emph{ICCV}, 2017.

\bibitem{dpc2019}
T.~Han, W.~Xie, and A.~Zisserman, ``Video representation learning by dense predictive coding,'' in \emph{ICCV}, 2019.

\bibitem{vcop2019}
D.~Xu, J.~Xiao, Z.~Zhao, J.~Shao, D.~Xie, and Y.~Zhuang, ``Self-supervised spatiotemporal learning via video clip order prediction,'' in \emph{CVPR}, 2019.

\bibitem{pacepred2020}
J.~Wang, J.~Jiao, and Y.-H. Liu, ``Self-supervised video representation learning by pace prediction,'' \emph{arXiv}, 2020.

\bibitem{vtdl2020}
J.~Wang, Y.~Lin, A.~J. Ma, and P.~C. Yuen, ``Self-supervised temporal discriminative learning for video representation learning,'' \emph{arXiv}, 2020.

\bibitem{prp2020}
Y.~Yao, C.~Liu, D.~Luo, Y.~Zhou, and Q.~Ye, ``Video playback rate perception for self-supervised spatio-temporal representation learning,'' in \emph{CVPR}, 2020.

\bibitem{vcp2020}
D.~Luo, C.~Liu, Y.~Zhou, D.~Yang, C.~Ma, Q.~Ye, and W.~Wang, ``Video cloze procedure for self-supervised spatio-temporal learning,'' \emph{arXiv}, 2020.

\bibitem{dsm2020}
J.~Wang, Y.~Gao, K.~Li, X.~Jiang, X.~Guo, R.~Ji, and X.~Sun, ``Enhancing unsupervised video representation learning by decoupling the scene and the motion,'' \emph{arXiv}, 2020.

\bibitem{ji20123d}
S.~Ji, W.~Xu, M.~Yang, and K.~Yu, ``3d convolutional neural networks for human action recognition,'' \emph{IEEE TPAMI}, 2012.

\bibitem{tran2015learning}
D.~Tran, L.~Bourdev, R.~Fergus, L.~Torresani, and M.~Paluri, ``Learning spatiotemporal features with 3d convolutional networks,'' in \emph{ICCV}, 2015.

\bibitem{feichtenhofer2020x3d}
C.~Feichtenhofer, ``X3d: Expanding architectures for efficient video recognition,'' in \emph{CVPR}, 2020.

\bibitem{feichtenhofer2019slowfast}
C.~Feichtenhofer, H.~Fan, J.~Malik, and K.~He, ``Slowfast networks for video recognition,'' in \emph{ICCV}, 2019.

\bibitem{wang2019fast}
S.~Wang, H.~Lu, and Z.~Deng, ``Fast object detection in compressed video,'' in \emph{ICCV}, 2019.

\bibitem{sun2019deep}
K.~Sun, B.~Xiao, D.~Liu, and J.~Wang, ``Deep high-resolution representation learning for human pose estimation,'' in \emph{CVPR}, 2019.

\bibitem{liu2016spatio}
J.~Liu, A.~Shahroudy, D.~Xu, and G.~Wang, ``Spatio-temporal lstm with trust gates for 3d human action recognition,'' in \emph{ECCV}.\hskip 1em plus 0.5em minus 0.4em\relax Springer, 2016.

\bibitem{zhang2017view}
P.~Zhang, C.~Lan, J.~Xing, W.~Zeng, J.~Xue, and N.~Zheng, ``View adaptive recurrent neural networks for high performance human action recognition from skeleton data,'' in \emph{ICCV}, 2017.

\bibitem{shi2019two}
L.~Shi, Y.~Zhang, J.~Cheng, and H.~Lu, ``Two-stream adaptive graph convolutional networks for skeleton-based action recognition,'' in \emph{CVPR}, 2019.

\bibitem{yan2018spatial}
S.~Yan, Y.~Xiong, and D.~Lin, ``Spatial temporal graph convolutional networks for skeleton-based action recognition,'' in \emph{AAAI}, 2018.

\bibitem{caetano2019skeleton}
C.~Caetano, F.~Br{\'e}mond, and W.~R. Schwartz, ``Skeleton image representation for 3d action recognition based on tree structure and reference joints,'' in \emph{2019 SIBGRAPI}.\hskip 1em plus 0.5em minus 0.4em\relax IEEE, 2019.

\bibitem{caetano2019skelemotion}
C.~Caetano, J.~Sena, F.~Br{\'e}mond, J.~A. Dos~Santos, and W.~R. Schwartz, ``Skelemotion: A new representation of skeleton joint sequences based on motion information for 3d action recognition,'' in \emph{2019 AVSS}.\hskip 1em plus 0.5em minus 0.4em\relax IEEE, 2019.

\bibitem{choutas2018potion}
V.~Choutas, P.~Weinzaepfel, J.~Revaud, and C.~Schmid, ``Potion: Pose motion representation for action recognition,'' in \emph{CVPR}, 2018.

\bibitem{das2021vpn++}
S.~Das, R.~Dai, D.~Yang, and F.~Bremond, ``Vpn++: Rethinking video-pose embeddings for understanding activities of daily living,'' \emph{IEEE TPAMI}, 2021.

\bibitem{das2020vpn}
S.~Das, S.~Sharma, R.~Dai, F.~Bremond, and M.~Thonnat, ``Vpn: Learning video-pose embedding for activities of daily living,'' in \emph{ECCV}.\hskip 1em plus 0.5em minus 0.4em\relax Springer, 2020.

\bibitem{gabeur2020multi}
V.~Gabeur, C.~Sun, K.~Alahari, and C.~Schmid, ``Multi-modal transformer for video retrieval,'' in \emph{ECCV}.\hskip 1em plus 0.5em minus 0.4em\relax Springer, 2020.

\bibitem{zhang2018cross}
B.~Zhang, H.~Hu, and F.~Sha, ``Cross-modal and hierarchical modeling of video and text,'' in \emph{ECCV}, 2018.

\bibitem{han2020memory}
T.~Han, W.~Xie, and A.~Zisserman, ``Memory-augmented dense predictive coding for video representation learning,'' in \emph{ECCV}.\hskip 1em plus 0.5em minus 0.4em\relax Springer, 2020.

\bibitem{hochreiter1997lstm}
S.~Hochreiter and J.~Schmidhuber, ``Long short-term memory,'' \emph{Neural computation}, 1997.

\bibitem{cho2014gru}
K.~Cho, B.~Van~Merri{\"e}nboer, C.~Gulcehre, D.~Bahdanau, F.~Bougares, H.~Schwenk, and Y.~Bengio, ``Learning phrase representations using rnn encoder-decoder for statistical machine translation,'' \emph{arXiv preprint arXiv:1406.1078}, 2014.

\bibitem{shen2021osgan}
C.~Shen, Y.~Yin, X.~Wang, X.~Li, J.~Song, and M.~Song, ``Training generative adversarial networks in one stage,'' in \emph{CVPR}, 2021.

\bibitem{ioffe2015batch}
S.~Ioffe and C.~Szegedy, ``Batch normalization: Accelerating deep network training by reducing internal covariate shift,'' in \emph{ICML}.\hskip 1em plus 0.5em minus 0.4em\relax PMLR, 2015.

\bibitem{clark2019dvdgan}
A.~Clark, J.~Donahue, and K.~Simonyan, ``Adversarial video generation on complex datasets,'' \emph{arXiv preprint arXiv:1907.06571}, 2019.

\bibitem{li2020moflowgan}
W.~Li, Z.~Yuan, X.~Fang, and C.~Wang, ``Moflowgan: Video generation with flow guidance,'' in \emph{2020 ICME}.\hskip 1em plus 0.5em minus 0.4em\relax IEEE, 2020.

\bibitem{kendall2018mtl}
A.~Kendall, Y.~Gal, and R.~Cipolla, ``Multi-task learning using uncertainty to weigh losses for scene geometry and semantics,'' in \emph{CVPR}, 2018.

\bibitem{caron2018deep}
M.~Caron, P.~Bojanowski, A.~Joulin, and M.~Douze, ``Deep clustering for unsupervised learning of visual features,'' in \emph{ECCV}, 2018.

\bibitem{wang2020self}
J.~Wang, J.~Jiao, and Y.-H. Liu, ``Self-supervised video representation learning by pace prediction,'' in \emph{ECCV}.\hskip 1em plus 0.5em minus 0.4em\relax Springer, 2020.

\bibitem{lee2017unsupervised}
H.-Y. Lee, J.-B. Huang, M.~Singh, and M.-H. Yang, ``Unsupervised representation learning by sorting sequences,'' in \emph{ICCV}, 2017.

\bibitem{huo2020self}
Y.~Huo, M.~Ding, H.~Lu, Z.~Lu, T.~Xiang, J.-R. Wen, Z.~Huang, J.~Jiang, S.~Zhang, M.~Tang \emph{et~al.}, ``Self-supervised video representation learning with constrained spatiotemporal jigsaw,'' \emph{IJCAI 2021}, 2020.

\bibitem{ahsan2019video}
U.~Ahsan, R.~Madhok, and I.~Essa, ``Video jigsaw: Unsupervised learning of spatiotemporal context for video action recognition,'' in \emph{2019 WACV}.\hskip 1em plus 0.5em minus 0.4em\relax IEEE, 2019.

\bibitem{chen2020simple}
T.~Chen, S.~Kornblith, M.~Norouzi, and G.~Hinton, ``A simple framework for contrastive learning of visual representations,'' in \emph{ICML}, 2020.

\bibitem{he2020momentum}
K.~He, H.~Fan, Y.~Wu, S.~Xie, and R.~Girshick, ``Momentum contrast for unsupervised visual representation learning,'' in \emph{CVPR}, 2020.

\bibitem{henaff2020data}
O.~Henaff, ``Data-efficient image recognition with contrastive predictive coding,'' in \emph{ICML}.\hskip 1em plus 0.5em minus 0.4em\relax PMLR, 2020.

\bibitem{hjelm2018learning}
R.~D. Hjelm, A.~Fedorov, S.~Lavoie-Marchildon, K.~Grewal, P.~Bachman, A.~Trischler, and Y.~Bengio, ``Learning deep representations by mutual information estimation and maximization,'' \emph{arXiv preprint arXiv:1808.06670}, 2018.

\bibitem{tian2020contrastive}
Y.~Tian, D.~Krishnan, and P.~Isola, ``Contrastive multiview coding,'' in \emph{ECCV}.\hskip 1em plus 0.5em minus 0.4em\relax Springer, 2020.

\bibitem{feichtenhofer2021large}
C.~Feichtenhofer, H.~Fan, B.~Xiong, R.~Girshick, and K.~He, ``A large-scale study on unsupervised spatiotemporal representation learning,'' in \emph{CVPR}, 2021.

\bibitem{asano2020labelling}
Y.~Asano, M.~Patrick, C.~Rupprecht, and A.~Vedaldi, ``Labelling unlabelled videos from scratch with multi-modal self-supervision,'' in \emph{NeurIPS}, 2020.

\bibitem{recasens2021broaden}
A.~Recasens, P.~Luc, J.-B. Alayrac, L.~Wang, F.~Strub, C.~Tallec, M.~Malinowski, V.~P{\u{a}}tr{\u{a}}ucean, F.~Altch{\'e}, M.~Valko \emph{et~al.}, ``Broaden your views for self-supervised video learning,'' in \emph{ICCV}, 2021.

\bibitem{behrmann2021long}
N.~Behrmann, M.~Fayyaz, J.~Gall, and M.~Noroozi, ``Long short view feature decomposition via contrastive video representation learning,'' in \emph{ICCV}, 2021.

\bibitem{dave2022tclr}
I.~Dave, R.~Gupta, M.~N. Rizve, and M.~Shah, ``Tclr: Temporal contrastive learning for video representation,'' \emph{CVIU}, 2022.

\bibitem{yu2020semantic}
Y.~Yu, T.~Xia, H.~Wang, J.~Feng, and Y.~Li, ``Semantic-aware spatio-temporal app usage representation via graph convolutional network,'' \emph{Proceedings of the ACM on Interactive, Mobile, Wearable and Ubiquitous Technologies}, 2020.

\bibitem{fragkiadaki2015recurrent}
K.~Fragkiadaki, S.~Levine, P.~Felsen, and J.~Malik, ``Recurrent network models for human dynamics,'' in \emph{ICCV}, 2015.

\bibitem{gulrajani2017improved}
I.~Gulrajani, F.~Ahmed, M.~Arjovsky, V.~Dumoulin, and A.~C. Courville, ``Improved training of wasserstein gans,'' in \emph{NeurIPS}, 2017.

\bibitem{martinez2017human}
J.~Martinez, M.~J. Black, and J.~Romero, ``On human motion prediction using recurrent neural networks,'' in \emph{CVPR}, 2017.

\bibitem{walker2017pose}
J.~Walker, K.~Marino, A.~Gupta, and M.~Hebert, ``The pose knows: Video forecasting by generating pose futures,'' in \emph{ICCV}, 2017.

\bibitem{vondrick2016vgan}
C.~Vondrick, H.~Pirsiavash, and A.~Torralba, ``Generating videos with scene dynamics,'' in \emph{NeurIPS}, 2016.

\bibitem{saito2017tgan}
M.~Saito, E.~Matsumoto, and S.~Saito, ``Temporal generative adversarial nets with singular value clipping,'' in \emph{ICCV}, 2017.

\bibitem{saito2020tganv2}
M.~Saito, S.~Saito, M.~Koyama, and S.~Kobayashi, ``Train sparsely, generate densely: Memory-efficient unsupervised training of high-resolution temporal gan,'' \emph{IJCV}, 2020.

\bibitem{fushishita2020long}
N.~Fushishita, A.~Tejero-de Pablos, Y.~Mukuta, and T.~Harada, ``Long-term human video generation of multiple futures using poses,'' in \emph{ECCV}.\hskip 1em plus 0.5em minus 0.4em\relax Springer, 2020.

\bibitem{cao2017realtime}
Z.~Cao, T.~Simon, S.-E. Wei, and Y.~Sheikh, ``Realtime multi-person 2d pose estimation using part affinity fields,'' in \emph{CVPR}, 2017.

\bibitem{tran2018closer}
D.~Tran, H.~Wang, L.~Torresani, J.~Ray, Y.~LeCun, and M.~Paluri, ``A closer look at spatiotemporal convolutions for action recognition,'' in \emph{CVPR}, 2018.

\bibitem{tran2019video}
D.~Tran, H.~Wang, L.~Torresani, and M.~Feiszli, ``Video classification with channel-separated convolutional networks,'' in \emph{ICCV}, 2019.

\bibitem{carreira2017quo}
J.~Carreira and A.~Zisserman, ``Quo vadis, action recognition? a new model and the kinetics dataset,'' in \emph{CVPR}, 2017.

\bibitem{mirza2014conditional}
M.~Mirza and S.~Osindero, ``Conditional generative adversarial nets,'' \emph{arXiv preprint arXiv:1411.1784}, 2014.

\bibitem{goodfellow2020generative}
I.~Goodfellow, J.~Pouget-Abadie, M.~Mirza, B.~Xu, D.~Warde-Farley, S.~Ozair, A.~Courville, and Y.~Bengio, ``Generative adversarial networks,'' \emph{Communications of the ACM}, 2020.

\bibitem{coskun2022goca}
H.~Coskun, A.~Zareian, J.~L. Moore, F.~Tombari, and C.~Wang, ``Goca: Guided online cluster assignment for self-supervised video representation learning,'' in \emph{ECCV}.\hskip 1em plus 0.5em minus 0.4em\relax Springer, 2022.

\bibitem{guo2022cross}
S.~Guo, Z.~Xiong, Y.~Zhong, L.~Wang, X.~Guo, B.~Han, and W.~Huang, ``Cross-architecture self-supervised video representation learning,'' in \emph{CVPR}, 2022.

\bibitem{tong2022videomae}
Z.~Tong, Y.~Song, J.~Wang, and L.~Wang, ``Videomae: Masked autoencoders are data-efficient learners for self-supervised video pre-training,'' in \emph{NeurIPS}, 2022.

\bibitem{chenempirical}
X.~Chen, S.~Xie, and K.~He, ``An empirical study of training self-supervised vision transformers. in 2021 ieee,'' in \emph{ICCV}.

\bibitem{dosovitskiy2020image}
A.~Dosovitskiy, L.~Beyer, A.~Kolesnikov, D.~Weissenborn, X.~Zhai, T.~Unterthiner, M.~Dehghani, M.~Minderer, G.~Heigold, S.~Gelly \emph{et~al.}, ``An image is worth 16x16 words: Transformers for image recognition at scale,'' \emph{arXiv preprint arXiv:2010.11929}, 2020.

\bibitem{bertasius2021space}
G.~Bertasius, H.~Wang, and L.~Torresani, ``Is space-time attention all you need for video understanding?'' in \emph{ICML}, 2021.

\bibitem{bulat2021space}
A.~Bulat, J.~M. Perez~Rua, S.~Sudhakaran, B.~Martinez, and G.~Tzimiropoulos, ``Space-time mixing attention for video transformer,'' in \emph{NeurIPS}, 2021.

\bibitem{caron2021emerging}
M.~Caron, H.~Touvron, I.~Misra, H.~J{\'e}gou, J.~Mairal, P.~Bojanowski, and A.~Joulin, ``Emerging properties in self-supervised vision transformers,'' in \emph{ICCV}, 2021.

\bibitem{wei2022masked}
C.~Wei, H.~Fan, S.~Xie, C.-Y. Wu, A.~Yuille, and C.~Feichtenhofer, ``Masked feature prediction for self-supervised visual pre-training,'' in \emph{CVPR}, 2022.

\bibitem{qian2021spatiotemporal}
R.~Qian, T.~Meng, B.~Gong, M.-H. Yang, H.~Wang, S.~Belongie, and Y.~Cui, ``Spatiotemporal contrastive video representation learning,'' in \emph{CVPR}, 2021.

\bibitem{3drotnet2018}
L.~Jing, X.~Yang, J.~Liu, and Y.~Tian, ``Self-supervised spatiotemporal feature learning via video rotation prediction,'' \emph{arXiv}, 2018.

\bibitem{stpuzzle2019}
D.~Kim, D.~Cho, and I.~S. Kweon, ``Self-supervised video representation learning with space-time cubic puzzles,'' in \emph{AAAI}, 2019.

\bibitem{cbt2019}
C.~Sun, F.~Baradel, K.~Murphy, and C.~Schmid, ``Learning video representations using contrastive bidirectional transformer,'' \emph{arXiv}, 2019.

\bibitem{speednet2020}
S.~Benaim, A.~Ephrat, O.~Lang, I.~Mosseri, W.~T. Freeman, M.~Rubinstein, M.~Irani, and T.~Dekel, ``Speednet: Learning the speediness in videos,'' in \emph{CVPR}, 2020.

\bibitem{memorydpc2020}
T.~Han, W.~Xie, and A.~Zisserman, ``Memory-augmented dense predictive coding for video representation learning,'' \emph{arXiv}, 2020.

\bibitem{vthcl2020}
C.~Yang, Y.~Xu, B.~Dai, and B.~Zhou, ``Video representation learning with visual tempo consistency,'' \emph{arXiv}, 2020.

\bibitem{jigsaw2016}
M.~Noroozi and P.~Favaro, ``Unsupervised learning of visual representations by solving jigsaw puzzles,'' in \emph{ECCV}.\hskip 1em plus 0.5em minus 0.4em\relax Springer, 2016.

\bibitem{buchler2018}
U.~Buchler, B.~Brattoli, and B.~Ommer, ``Improving spatiotemporal self-supervision by deep reinforcement learning,'' in \emph{ECCV}, 2018.

\bibitem{xdc2019}
H.~Alwassel, D.~Mahajan, L.~Torresani, B.~Ghanem, and D.~Tran, ``Self-supervised learning by cross-modal audio-video clustering,'' \emph{arXiv}, 2019.

\bibitem{atvs2018}
B.~Korbar, D.~Tran, and L.~Torresani, ``Cooperative learning of audio and video models from self-supervised synchronization,'' in \emph{NeurIPS}, 2018.

\bibitem{gdt2020}
M.~Patrick, Y.~M. Asano, R.~Fong, J.~F. Henriques, G.~Zweig, and A.~Vedaldi, ``Multi-modal self-supervision from generalized data transformations,'' \emph{arXiv}, 2020.

\bibitem{milnce2020}
A.~Miech, J.-B. Alayrac, L.~Smaira, I.~Laptev, J.~Sivic, and A.~Zisserman, ``End-to-end learning of visual representations from uncurated instructional videos,'' in \emph{CVPR}, 2020.

\bibitem{twostream2014}
K.~Simonyan and A.~Zisserman, ``Two-stream convolutional networks for action recognition in videos,'' in \emph{NeurIPS}, 2014.

\bibitem{disentangle2018}
Y.~Zhao, Y.~Xiong, and D.~Lin, ``Recognize actions by disentangling components of dynamics,'' in \emph{CVPR}, 2018.

\bibitem{diving48_2018}
Y.~Li, Y.~Li, and N.~Vasconcelos, ``Resound: Towards action recognition without representation bias,'' in \emph{Proceedings of the ECCV}, 2018.

\bibitem{tvl1_2007}
C.~Zach, T.~Pock, and H.~Bischof, ``A duality based approach for realtime tv-l 1 optical flow,'' in \emph{Joint pattern recognition symposium}.\hskip 1em plus 0.5em minus 0.4em\relax Springer, 2007.

\bibitem{tvl1_2009}
F.~Steinbr{\"u}cker, T.~Pock, and D.~Cremers, ``Large displacement optical flow computation withoutwarping,'' in \emph{ICCV}.\hskip 1em plus 0.5em minus 0.4em\relax IEEE, 2009.

\bibitem{highacc2004}
T.~Brox, A.~Bruhn, N.~Papenberg, and J.~Weickert, ``High accuracy optical flow estimation based on a theory for warping,'' in \emph{ECCV}.\hskip 1em plus 0.5em minus 0.4em\relax Springer, 2004.

\bibitem{densetraj2011}
\BIBentryALTinterwordspacing
H.~Wang, A.~Kl{\"a}ser, C.~Schmid, and C.-L. Liu, ``{Action Recognition by Dense Trajectories},'' in \emph{IEEE Conference on Computer Vision \& Pattern Recognition}, Colorado Springs, United States, Jun. 2011, pp. 3169--3176. [Online]. Available: \url{http://hal.inria.fr/inria-00583818/en}
\BIBentrySTDinterwordspacing

\bibitem{videobert2019}
C.~Sun, A.~Myers, C.~Vondrick, K.~Murphy, and C.~Schmid, ``Videobert: A joint model for video and language representation learning,'' in \emph{ICCV}, 2019.

\bibitem{coviar2018}
C.-Y. Wu, M.~Zaheer, H.~Hu, R.~Manmatha, A.~J. Smola, and P.~Kr{\"a}henb{\"u}hl, ``Compressed video action recognition,'' in \emph{CVPR}, 2018.

\bibitem{nce2010}
M.~Gutmann and A.~Hyv{\"a}rinen, ``Noise-contrastive estimation: A new estimation principle for unnormalized statistical models,'' in \emph{Proceedings of the Thirteenth International Conference on Artificial Intelligence and Statistics}, 2010.

\bibitem{infonce2018}
A.~v.~d. Oord, Y.~Li, and O.~Vinyals, ``Representation learning with contrastive predictive coding,'' \emph{arXiv}, 2018.

\bibitem{li2019faceshifter}
L.~Li, J.~Bao, H.~Yang, D.~Chen, and F.~Wen, ``Faceshifter: Towards high fidelity and occlusion aware face swapping,'' \emph{arXiv preprint arXiv:1912.13457}, 2019.

\bibitem{huang2017arbitrary}
X.~Huang and S.~Belongie, ``Arbitrary style transfer in real-time with adaptive instance normalization,'' in \emph{ICCV}, 2017.

\bibitem{ucf101}
K.~Soomro, A.~R. Zamir, and M.~Shah, ``Ucf101: A dataset of 101 human actions classes from videos in the wild,'' \emph{arXiv}, 2012.

\bibitem{hmdb51}
H.~Kuehne, H.~Jhuang, E.~Garrote, T.~Poggio, and T.~Serre, ``Hmdb: a large video database for human motion recognition,'' in \emph{ICCV}.\hskip 1em plus 0.5em minus 0.4em\relax IEEE, 2011.

\bibitem{simclr2020}
T.~Chen, S.~Kornblith, M.~Norouzi, and G.~Hinton, ``A simple framework for contrastive learning of visual representations,'' \emph{arXiv}, 2020.

\bibitem{simclrv2_2020}
T.~Chen, S.~Kornblith, K.~Swersky, M.~Norouzi, and G.~Hinton, ``Big self-supervised models are strong semi-supervised learners,'' \emph{arXiv}, 2020.

\bibitem{moco2019}
K.~He, H.~Fan, Y.~Wu, S.~Xie, and R.~Girshick, ``Momentum contrast for unsupervised visual representation learning,'' \emph{arXiv}, 2019.

\bibitem{mocov2_2020}
X.~Chen, H.~Fan, R.~Girshick, and K.~He, ``Improved baselines with momentum contrastive learning,'' \emph{arXiv}, 2020.

\bibitem{avslowfast2020}
F.~Xiao, Y.~J. Lee, K.~Grauman, J.~Malik, and C.~Feichtenhofer, ``Audiovisual slowfast networks for video recognition,'' \emph{arXiv}, 2020.

\bibitem{elo2020}
A.~Piergiovanni, A.~Angelova, and M.~S. Ryoo, ``Evolving losses for unsupervised video representation learning,'' in \emph{CVPR}, 2020.

\bibitem{dmc2019}
Z.~Shou, X.~Lin, Y.~Kalantidis, L.~Sevilla-Lara, M.~Rohrbach, S.-F. Chang, and Z.~Yan, ``Dmc-net: Generating discriminative motion cues for fast compressed video action recognition,'' in \emph{CVPR}, 2019.

\bibitem{mv2016}
B.~Zhang, L.~Wang, Z.~Wang, Y.~Qiao, and H.~Wang, ``Real-time action recognition with enhanced motion vector cnns,'' in \emph{CVPR}, 2016.

\bibitem{mv2018}
------, ``Real-time action recognition with deeply transferred motion vector cnns,'' \emph{IEEE TIP}, 2018.

\bibitem{compressed_detection2019}
S.~Wang, H.~Lu, and Z.~Deng, ``Fast object detection in compressed video,'' in \emph{ICCV}, 2019.

\bibitem{video_compression_1991}
D.~Le~Gall, ``Mpeg: A video compression standard for multimedia applications,'' \emph{Communications of the ACM}, 1991.

\bibitem{human_motion_prediction_2015}
K.~Fragkiadaki, S.~Levine, P.~Felsen, and J.~Malik, ``Recurrent network models for human dynamics,'' in \emph{ICCV}, 2015.

\bibitem{human_motion_prediction_2016}
A.~Jain, A.~R. Zamir, S.~Savarese, and A.~Saxena, ``Structural-rnn: Deep learning on spatio-temporal graphs,'' in \emph{CVPR}, 2016.

\bibitem{human_motion_prediction_2017}
J.~Martinez, M.~J. Black, and J.~Romero, ``On human motion prediction using recurrent neural networks,'' in \emph{CVPR}, 2017.

\bibitem{human_motion_prediction_2019}
W.~Mao, M.~Liu, M.~Salzmann, and H.~Li, ``Learning trajectory dependencies for human motion prediction,'' in \emph{ICCV}, 2019.

\bibitem{traj_prediction_2020}
L.~Fang, Q.~Jiang, J.~Shi, and B.~Zhou, ``Tpnet: Trajectory proposal network for motion prediction,'' in \emph{CVPR}, 2020.

\bibitem{traj_prediction_2018}
T.~Yagi, K.~Mangalam, R.~Yonetani, and Y.~Sato, ``Future person localization in first-person videos,'' in \emph{CVPR}, 2018.

\bibitem{mp_rnn2018}
N.~Djuric, V.~Radosavljevic, H.~Cui, T.~Nguyen, F.-C. Chou, T.-H. Lin, and J.~Schneider, ``Motion prediction of traffic actors for autonomous driving using deep convolutional networks,'' \emph{arXiv}, 2018.

\bibitem{mp_rnn2017}
D.~Lee, Y.~P. Kwon, S.~McMains, and J.~K. Hedrick, ``Convolution neural network-based lane change intention prediction of surrounding vehicles for acc,'' in \emph{2017 IEEE 20th International Conference on Intelligent Transportation Systems (ITSC)}.\hskip 1em plus 0.5em minus 0.4em\relax IEEE, 2017.

\bibitem{mp_transformer_2020}
E.~Aksan, P.~Cao, M.~Kaufmann, and O.~Hilliges, ``Attention, please: A spatio-temporal transformer for 3d human motion prediction,'' \emph{arXiv}, 2020.

\bibitem{mp_transformer_2020_2}
C.~Yu, X.~Ma, J.~Ren, H.~Zhao, and S.~Yi, ``Spatio-temporal graph transformer networks for pedestrian trajectory prediction,'' \emph{arXiv}, 2020.

\bibitem{transformer2017}
A.~Vaswani, N.~Shazeer, N.~Parmar, J.~Uszkoreit, L.~Jones, A.~N. Gomez, {\L}.~Kaiser, and I.~Polosukhin, ``Attention is all you need,'' in \emph{NeurIPS}, 2017.

\bibitem{swav2020}
M.~Caron, I.~Misra, J.~Mairal, P.~Goyal, P.~Bojanowski, and A.~Joulin, ``Unsupervised learning of visual features by contrasting cluster assignments,'' \emph{arXiv}, 2020.

\bibitem{patrick2021compositions}
M.~Patrick, Y.~M. Asano, P.~Kuznetsova, R.~Fong, J.~F. Henriques, G.~Zweig, and A.~Vedaldi, ``On compositions of transformations in contrastive self-supervised learning,'' in \emph{ICCV}, 2021.

\bibitem{sela2020}
Y.~M. Asano, C.~Rupprecht, and A.~Vedaldi, ``Self-labelling via simultaneous clustering and representation learning,'' in \emph{International Conference on Learning Representations (ICLR)}, 2020.

\bibitem{convgru2015}
N.~Ballas, L.~Yao, C.~Pal, and A.~Courville, ``Delving deeper into convolutional networks for learning video representations,'' \emph{arXiv}, 2015.

\bibitem{kay2017kinetics}
W.~Kay, J.~Carreira, K.~Simonyan, B.~Zhang, C.~Hillier, S.~Vijayanarasimhan, F.~Viola, T.~Green, T.~Back, P.~Natsev \emph{et~al.}, ``The kinetics human action video dataset,'' \emph{arXiv preprint arXiv:1705.06950}, 2017.

\bibitem{c3d2015}
D.~Tran, L.~Bourdev, R.~Fergus, L.~Torresani, and M.~Paluri, ``Learning spatiotemporal features with 3d convolutional networks,'' in \emph{ICCV}, 2015.

\bibitem{resnet2p1d_2018}
D.~Tran, H.~Wang, L.~Torresani, J.~Ray, Y.~LeCun, and M.~Paluri, ``A closer look at spatiotemporal convolutions for action recognition,'' in \emph{CVPR}, 2018.

\bibitem{resnet2016}
K.~He, X.~Zhang, S.~Ren, and J.~Sun, ``Deep residual learning for image recognition,'' in \emph{CVPR}, 2016.

\bibitem{s3d2018}
S.~Xie, C.~Sun, J.~Huang, Z.~Tu, and K.~Murphy, ``Rethinking spatiotemporal feature learning: Speed-accuracy trade-offs in video classification,'' in \emph{ECCV}, 2018.

\bibitem{ding2022motion}
S.~Ding, M.~Li, T.~Yang, R.~Qian, H.~Xu, Q.~Chen, J.~Wang, and H.~Xiong, ``Motion-aware contrastive video representation learning via foreground-background merging,'' in \emph{CVPR}, 2022.

\bibitem{yao2020playback}
Y.~Yao, C.~Liu, D.~Luo, Y.~Zhou, and Q.~Ye, ``Video playback rate perception for self-supervised spatio-temporal representation learning,'' in \emph{CVPR}, 2020.

\bibitem{han2020self}
T.~Han, W.~Xie, and A.~Zisserman, ``Self-supervised co-training for video representation learning,'' in \emph{NeurIPS}, 2020.

\bibitem{qian2021enhancing}
R.~Qian, Y.~Li, H.~Liu, J.~See, S.~Ding, X.~Liu, D.~Li, and W.~Lin, ``Enhancing self-supervised video representation learning via multi-level feature optimization,'' in \emph{ICCV}, 2021.

\bibitem{wang2021self}
J.~Wang, J.~Jiao, L.~Bao, S.~He, W.~Liu, and Y.-H. Liu, ``Self-supervised video representation learning by uncovering spatio-temporal statistics,'' \emph{IEEE TPAMI}, 2021.

\bibitem{lin2021self}
Y.~Lin, X.~Guo, and Y.~Lu, ``Self-supervised video representation learning with meta-contrastive network,'' in \emph{ICCV}, 2021.

\bibitem{linear_lr_2017}
P.~Goyal, P.~Doll{\'a}r, R.~Girshick, P.~Noordhuis, L.~Wesolowski, A.~Kyrola, A.~Tulloch, Y.~Jia, and K.~He, ``Accurate, large minibatch sgd: Training imagenet in 1 hour,'' \emph{arXiv}, 2017.

\bibitem{huang2021self}
L.~Huang, Y.~Liu, B.~Wang, P.~Pan, Y.~Xu, and R.~Jin, ``Self-supervised video representation learning by context and motion decoupling,'' in \emph{CVPR}, 2021.

\bibitem{diba2021vi2clr}
A.~Diba, V.~Sharma, R.~Safdari, D.~Lotfi, S.~Sarfraz, R.~Stiefelhagen, and L.~Van~Gool, ``Vi2clr: Video and image for visual contrastive learning of representation,'' in \emph{ICCV}, 2021.

\bibitem{diba2019dynamonet}
A.~Diba, V.~Sharma, L.~V. Gool, and R.~Stiefelhagen, ``Dynamonet: Dynamic action and motion network,'' in \emph{ICCV}, 2019.

\bibitem{liang2017dual}
X.~Liang, L.~Lee, W.~Dai, and E.~P. Xing, ``Dual motion gan for future-flow embedded video prediction,'' in \emph{ICCV}, 2017.

\bibitem{ghadekar2023semi}
P.~Ghadekar, D.~Khanwelkar, N.~Soni, H.~More, J.~Rajani, and C.~Vaswani, ``A semi-supervised gan architecture for video classification,'' in \emph{AICAPS}.\hskip 1em plus 0.5em minus 0.4em\relax IEEE, 2023.

\bibitem{ranasinghe2022self}
K.~Ranasinghe, M.~Naseer, S.~Khan, F.~S. Khan, and M.~S. Ryoo, ``Self-supervised video transformer,'' in \emph{CVPR}, 2022.

\bibitem{hu2021contrast}
K.~Hu, J.~Shao, Y.~Liu, B.~Raj, M.~Savvides, and Z.~Shen, ``Contrast and order representations for video self-supervised learning,'' in \emph{ICCV}, 2021.

\bibitem{wang2024knowledge}
G.~Wang, Y.~Zhou, Z.~He, K.~Lu, Y.~Feng, Z.~Liu, and G.~Wang, ``Knowledge-guided pre-training and fine-tuning: Video representation learning for action recognition,'' \emph{Neurocomputing}, 2024.

\bibitem{lin2024tsgan}
W.~Lin, H.~Zeng, J.~Zhu, C.-H. Hsia, J.~Hou, and K.-K. Ma, ``Unsupervised video-based action recognition using two-stream generative adversarial network,'' \emph{Neural Computing and Applications}, 2024.

\bibitem{goyal2017ssv2}
R.~Goyal, S.~Ebrahimi~Kahou, V.~Michalski, J.~Materzynska, S.~Westphal, H.~Kim, V.~Haenel, I.~Fruend, P.~Yianilos, M.~Mueller-Freitag \emph{et~al.}, ``The" something something" video database for learning and evaluating visual common sense,'' in \emph{ICCV}, 2017.

\bibitem{kingma2013vae}
D.~P. Kingma, ``Auto-encoding variational bayes,'' \emph{arXiv preprint arXiv:1312.6114}, 2013.

\bibitem{ho2020diffusion}
J.~Ho, A.~Jain, and P.~Abbeel, ``Denoising diffusion probabilistic models,'' in \emph{NeurIPS}, 2020.

\bibitem{ge2022long}
S.~Ge, T.~Hayes, H.~Yang, X.~Yin, G.~Pang, D.~Jacobs, J.-B. Huang, and D.~Parikh, ``Long video generation with time-agnostic vqgan and time-sensitive transformer,'' in \emph{ECCV}.\hskip 1em plus 0.5em minus 0.4em\relax Springer, 2022.

\bibitem{villegas2022phenaki}
R.~Villegas, M.~Babaeizadeh, P.-J. Kindermans, H.~Moraldo, H.~Zhang, M.~T. Saffar, S.~Castro, J.~Kunze, and D.~Erhan, ``Phenaki: Variable length video generation from open domain textual descriptions,'' in \emph{ICLR}, 2022.

\bibitem{yan2021videogpt}
W.~Yan, Y.~Zhang, P.~Abbeel, and A.~Srinivas, ``Videogpt: Video generation using vq-vae and transformers,'' \emph{arXiv preprint arXiv:2104.10157}, 2021.

\bibitem{nichol2021improved}
A.~Q. Nichol and P.~Dhariwal, ``Improved denoising diffusion probabilistic models,'' in \emph{ICML}.\hskip 1em plus 0.5em minus 0.4em\relax PMLR, 2021.

\bibitem{ramesh2022hierarchical}
A.~Ramesh, P.~Dhariwal, A.~Nichol, C.~Chu, and M.~Chen, ``Hierarchical text-conditional image generation with clip latents.''

\bibitem{rombach2022high}
R.~Rombach, A.~Blattmann, D.~Lorenz, P.~Esser, and B.~Ommer, ``High-resolution image synthesis with latent diffusion models,'' in \emph{CVPR}, 2022.

\bibitem{saharia2022photorealistic}
C.~Saharia, W.~Chan, S.~Saxena, L.~Li, J.~Whang, E.~L. Denton, K.~Ghasemipour, R.~Gontijo~Lopes, B.~Karagol~Ayan, T.~Salimans \emph{et~al.}, ``Photorealistic text-to-image diffusion models with deep language understanding,'' in \emph{NeurIPS}, 2022.

\bibitem{ho2022video}
J.~Ho, T.~Salimans, A.~Gritsenko, W.~Chan, M.~Norouzi, and D.~J. Fleet, ``Video diffusion models,'' in \emph{NeurIPS}, 2022.

\bibitem{voleti2022mcvd}
V.~Voleti, A.~Jolicoeur-Martineau, and C.~Pal, ``Mcvd-masked conditional video diffusion for prediction, generation, and interpolation,'' in \emph{NeurIPS}, 2022.

\bibitem{blattmann2023align}
A.~Blattmann, R.~Rombach, H.~Ling, T.~Dockhorn, S.~W. Kim, S.~Fidler, and K.~Kreis, ``Align your latents: High-resolution video synthesis with latent diffusion models,'' in \emph{CVPR}, 2023.

\bibitem{ge2023preserve}
S.~Ge, S.~Nah, G.~Liu, T.~Poon, A.~Tao, B.~Catanzaro, D.~Jacobs, J.-B. Huang, M.-Y. Liu, and Y.~Balaji, ``Preserve your own correlation: A noise prior for video diffusion models,'' in \emph{ICCV}, 2023.

\bibitem{singer2022make}
U.~Singer, A.~Polyak, T.~Hayes, X.~Yin, J.~An, S.~Zhang, Q.~Hu, H.~Yang, O.~Ashual, O.~Gafni \emph{et~al.}, ``Make-a-video: Text-to-video generation without text-video data,'' \emph{arXiv preprint arXiv:2209.14792}, 2022.

\bibitem{luo2023videofusion}
Z.~Luo, D.~Chen, Y.~Zhang, Y.~Huang, L.~Wang, Y.~Shen, D.~Zhao, J.~Zhou, and T.~Tan, ``Videofusion: Decomposed diffusion models for high-quality video generation,'' \emph{arXiv preprint arXiv:2303.08320}, 2023.

\bibitem{zhao2024cv}
S.~Zhao, Y.~Zhang, X.~Cun, S.~Yang, M.~Niu, X.~Li, W.~Hu, and Y.~Shan, ``Cv-vae: A compatible video vae for latent generative video models,'' \emph{arXiv preprint arXiv:2405.20279}, 2024.

\bibitem{gupta2025photorealistic}
A.~Gupta, L.~Yu, K.~Sohn, X.~Gu, M.~Hahn, F.-F. Li, I.~Essa, L.~Jiang, and J.~Lezama, ``Photorealistic video generation with diffusion models,'' in \emph{ECCV}, 2025.

\end{thebibliography}
\bibliographystyle{IEEEtran}
 
\vspace{11pt}
\begin{IEEEbiography}[{\includegraphics[width=1in,height=1.25in,clip,keepaspectratio]{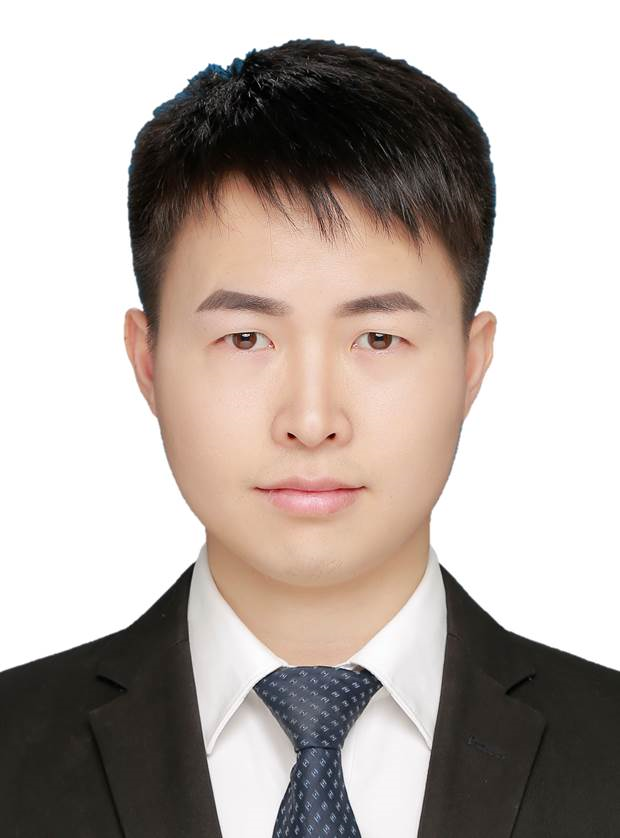}}]{Jie Zhang}
 (Member, IEEE) received the Ph.D. degree from the University of Chinese Academy of Sciences (CAS), Beijing, China. He is currently an Associate Professor with the Institute of Computing Technology, CAS. His research interests include computer vision, pattern recognition, machine learning, particularly include AI safety, face recognition, self-supervised learning, and domain generalization. He was successively selected for the Beijing Science and Technology Nova Program, the Youth Innovation Promotion Association of the Chinese Academy of Sciences, and the MSRA "Casting Star Project".
\end{IEEEbiography}

\begin{IEEEbiography}[{\includegraphics[width=1in,height=1.25in,clip,keepaspectratio]{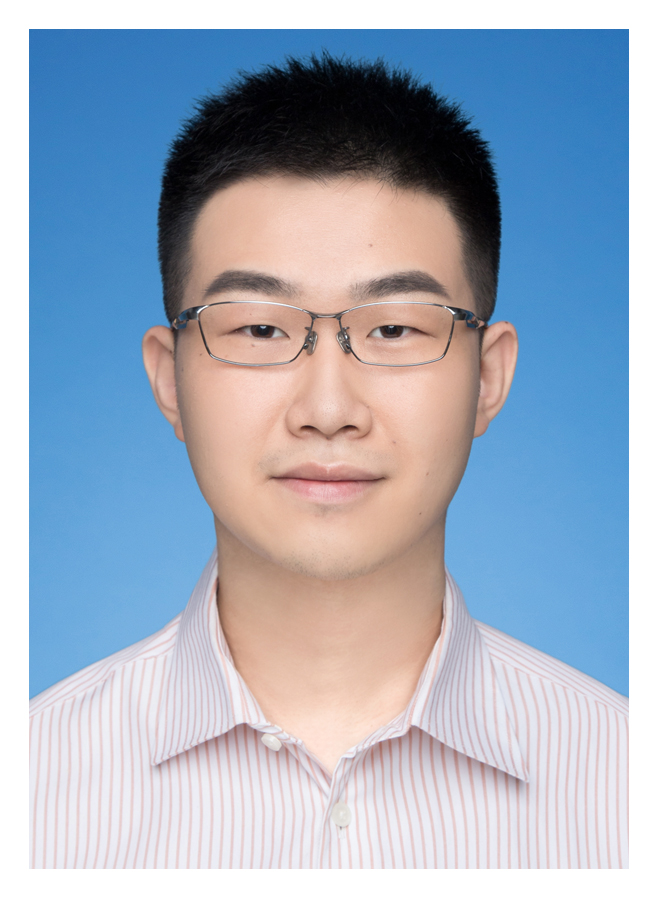}}]{Zhifan Wan}
eceived the BS degree from the University of Chinese Academy of Sciences Beijing, China, in 2021 and the M.S. degree from the Institute of Computing Technology (ICT), Chinese Academy of Sciences (CAS), in 2024. During the graduate studies, his research focused on video representation learning, self-supervised learning, action recognition, and action retrieval.
\end{IEEEbiography}

\begin{IEEEbiography}[{\includegraphics[width=1in,height=1.25in,clip,keepaspectratio]{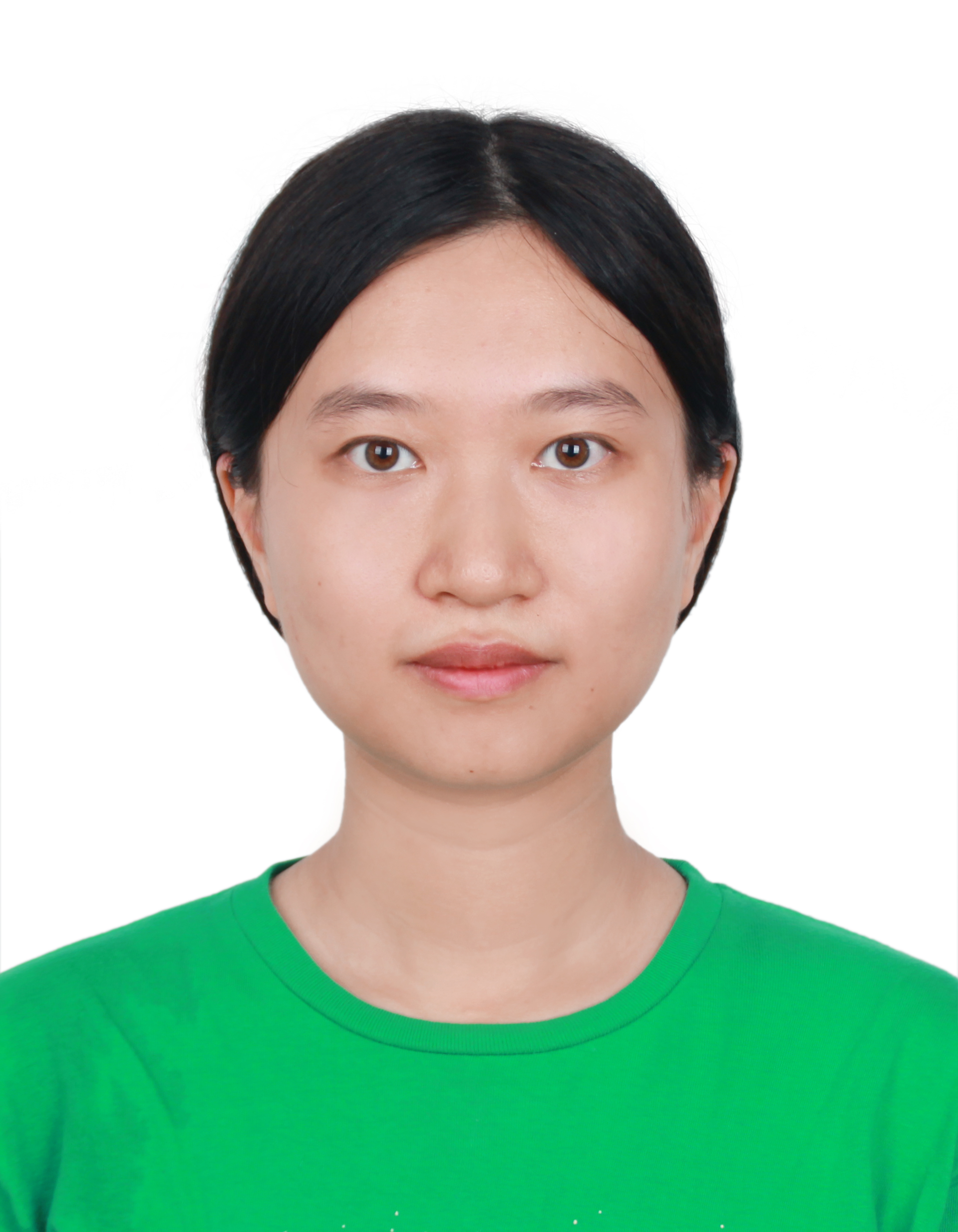}}]{Lanqing Hu} received the BS degree in software engineering from the Beijing University of Posts and Telecommunications, Beijing, China, in 2014 and the PhD degree in computer science from the Institute of Computing Technology (ICT), Chinese Academy of Sciences (CAS). She was associated with the Key Lab of Intelligent Information Processing, Institute of Computing Technology (ICT), Chinese Academy of Sciences (CAS), Beijing, China, and the University of Chinese Academy of Sciences (UCAS), Beijing, China. Her research interests cover computer vision and machine learning.
\end{IEEEbiography}

\begin{IEEEbiography}[{\includegraphics[width=1in,height=1.25in,clip,keepaspectratio]{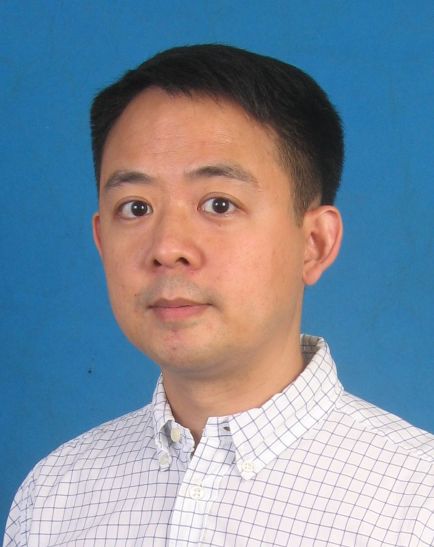}}]{Stephen Lin}
is a Senior Principal Research Manager in the Visual Computing Group of Microsoft Research Asia. He received his B.S.E. degree from Princeton University and his Ph.D. degree from the University of Michigan. He is on the editorial boards of the International Journal of Computer Vision and Transactions on Machine Learning Research. He has served as a program co-chair for International Conference on Computer Vision 2011, International Conference on 3D Vision 2020, and Computer Graphics International 2023. His research interests include computer vision, machine learning, and computer graphics.
\end{IEEEbiography}

\begin{IEEEbiography}[{\includegraphics[width=1in,height=1.25in,clip,keepaspectratio]{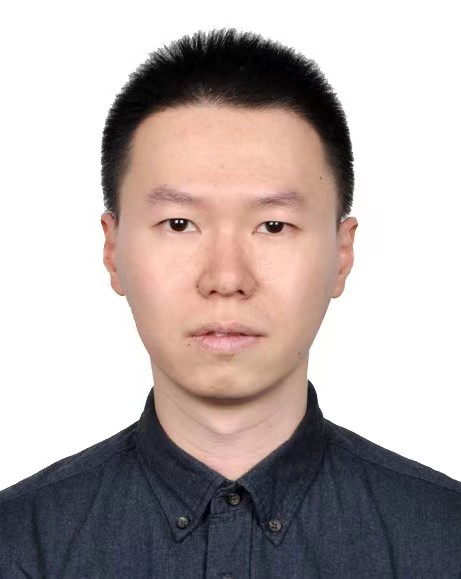}}]{Shuzhe Wu}
is currently affiliated with Beijing Huawei Digital Technologies Co., Ltd. He received the Ph.D. degree from the Institute of Computing Technology, Chinese Academy of Sciences, Beijing, China, in 2019. Dr. Wu’s research focuses on computer vision, including image/video understanding and generation, 3D reconstruction and neural rendering. He has published more than 10 refereed papers on journals and conferences, including CVPR, ECCV, IJCV, and served as a reviewer for CVPR, ICCV, ECCV, 3DV, TKDE, TKDD. He applied more than 10 patents.
\end{IEEEbiography}
\begin{IEEEbiography}[{\includegraphics[width=1in,height=1.25in,clip,keepaspectratio]{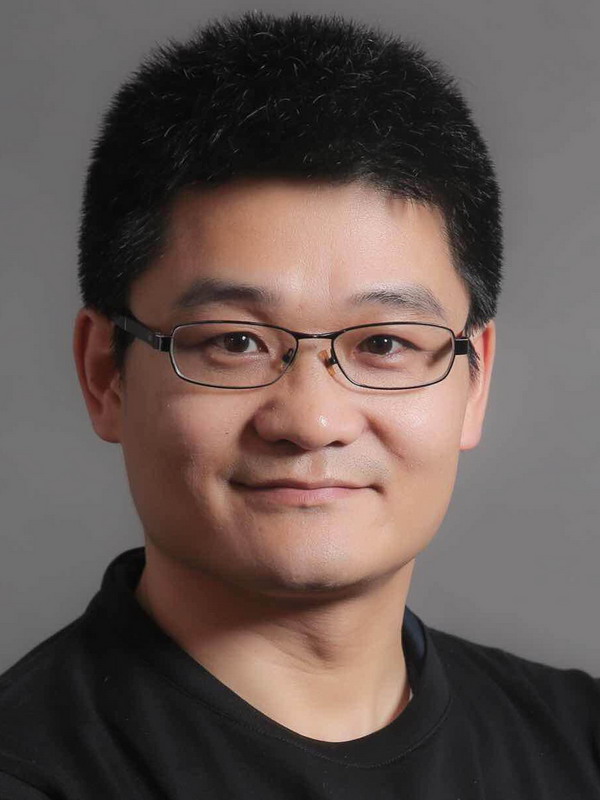}}]{Shiguang Shan}
(Fellow, IEEE) received the Ph.D. degree in computer science from the Institute of Computing Technology (ICT), Chinese Academy of Sciences (CAS), Beijing, China, in 2004. He has been a Full Professor with ICT since 2010, where he is currently the Director of the Key Laboratory of Intelligent Information Processing, CAS. His research interests include signal processing, computer vision, pattern recognition, and machine learning. He has published more than 300 articles in related areas. He served as the General Co-Chair for IEEE Face and Gesture Recognition 2023, the General Co-Chair for Asian Conference on Computer Vision (ACCV) 2022, and the Area Chair of many international conferences, including CVPR, ICCV, AAAI, IJCAI, ACCV, ICPR, and FG. He was/is an Associate Editors of several journals, including IEEE TRANSACTIONS ON IMAGE PROCESSING, Neurocomputing, CVIU, and PRL. He was a recipient of the China’s State Natural Science Award in 2015 and the China’s State S\&T Progress Award in 2005 for his research work.
\end{IEEEbiography}
\vspace{11pt}

\vfill

\end{document}